\documentclass[review]{elsarticle}
\usepackage[whole]{bxcjkjatype}

\usepackage{bm}

\usepackage{amssymb}
\usepackage{amsmath}
\usepackage{color}

\usepackage[linesnumbered,ruled,vlined]{algorithm2e}
\usepackage{algorithm2e,setspace}

\SetCommentSty{mycommfont}

\usepackage{booktabs}

\usepackage{graphicx}
\usepackage{floatrow}
\usepackage{subfig}

\usepackage{array}
\newcolumntype{C}[1]{>{\centering\arraybackslash}p{#1}}

\newcommand{\argmax}{\mathop{\rm arg~max}\limits}

\usepackage{lineno,hyperref}
\modulolinenumbers[5]

\begin{document}

\begin{frontmatter}

\title{Goal-Aware Generative Adversarial Imitation Learning from Imperfect Demonstration for Robotic Cloth Manipulation}

\author[naistaddress]{Yoshihisa Tsurumine\corref{cor1}}
\ead{tsurumine.yoshihisa@is.naist.jp}

\author[naistaddress]{Takamitsu Matsubara}
\ead{takam-m@is.naist.jp}

\cortext[cor1]{Corresponding author}

\address[naistaddress]{Division of Information Science, Graduate School of Science and Technology, Nara Institute of Science and Technology, 8916-5 Takayamacho, Ikoma, Nara, Japan}

\begin{abstract}
Generative Adversarial Imitation Learning (GAIL) can learn policies without explicitly defining the reward function from demonstrations. GAIL has the potential to learn policies with high-dimensional observations as input, e.g., images. By applying GAIL to a real robot, perhaps robot policies can be obtained for daily activities like washing, folding clothes, cooking, and cleaning. 
However, human demonstration data are often imperfect due to mistakes, which degrade the performance of the resulting policies. 
We address this issue by focusing on the following features: 1) many robotic tasks are goal-reaching tasks, and 2) labeling such goal states in demonstration data is relatively easy. 
With these in mind, this paper proposes Goal-Aware Generative Adversarial Imitation Learning (GA-GAIL), which trains a policy by introducing a second discriminator to distinguish the goal state in parallel with the first discriminator that indicates the demonstration data. This extends a standard GAIL framework to more robustly learn desirable policies even from imperfect demonstrations through a goal-state discriminator that promotes achieving the goal state.
Furthermore, GA-GAIL employs the Entropy-maximizing Deep P-Network (EDPN) as a generator, which considers both the smoothness and causal entropy in the policy update, to achieve stable policy learning from two discriminators. Our proposed method was successfully applied to two real-robotic cloth-manipulation tasks: turning a handkerchief over and folding clothes. We confirmed that it learns cloth-manipulation policies without task-specific reward function design. Video of the real experiments are available at \href{https://youtu.be/h_nII2ooUrE}{this URL}.
\end{abstract}

\begin{keyword}
\texttt generative adversarial imitation learning \sep robotic cloth manipulation \sep deep reinforcement learning
\MSC[2022] 00-01\sep  99-00
\end{keyword}

\end{frontmatter}


\section{Introduction}
Generative Adversarial Imitation Learning (GAIL) \cite{GAIL} can learn control policies using as input such high-dimensional observations as images. It has the potential to be applied for robots to conduct daily tasks involving many non-rigid objects, such as clothing, liquids, and food, that involve complex state transitions of flexible objects due to their interaction with the environment. 

Comparing GAIL and deep reinforcement learning (DRL), DRL requires the design of a reward function for each task to evaluate the learning policies from sensor information \cite{RASDPN, DBLP:conf/icra/Zhu0RLK19, pmlr-v100-nagabandi20a}. On the other hand, GAIL learns a discriminator between the generated and demonstration trajectories and evaluates the learning policies with a reward function based on the discriminator. Since GAIL does not require the design of a reward function for each task, it is suitable for daily tasks where the design of reward functions is complex.

However, when GAIL is applied to real-robot tasks using human demonstration data, the human demonstrations are often not perfect. Such imperfections might be caused by a lack of understanding or experience with the task. In particular, flexible-object manipulation tasks have complex state transitions due to the object's high deformability, so demonstrations are likely to contain mistakes. Such imperfect demonstrations considerably degrade the performance of the learned policies in GAIL. 

This paper addresses this issue by focusing on the following features: 1) many robotic tasks are goal-reaching, and 2) labeling such goal states in demonstration data is relatively easy. Therefore, this paper proposes Goal-Aware Generative Adversarial Imitation Learning (GA-GAIL), which trains policies by introducing a second discriminator that distinguishes the goal state in parallel with the first discriminator that indicates the demonstration data. This extends a standard GAIL framework to more robustly learn desirable policies even from imperfect demonstrations through a goal-state discriminator that promotes achieving the goal state. 

\begin{figure}
\begin{minipage}{0.5\textwidth}
\centering
\subfloat[NEXTAGE: dual-arm humanoid robot]{\includegraphics[width=6cm]{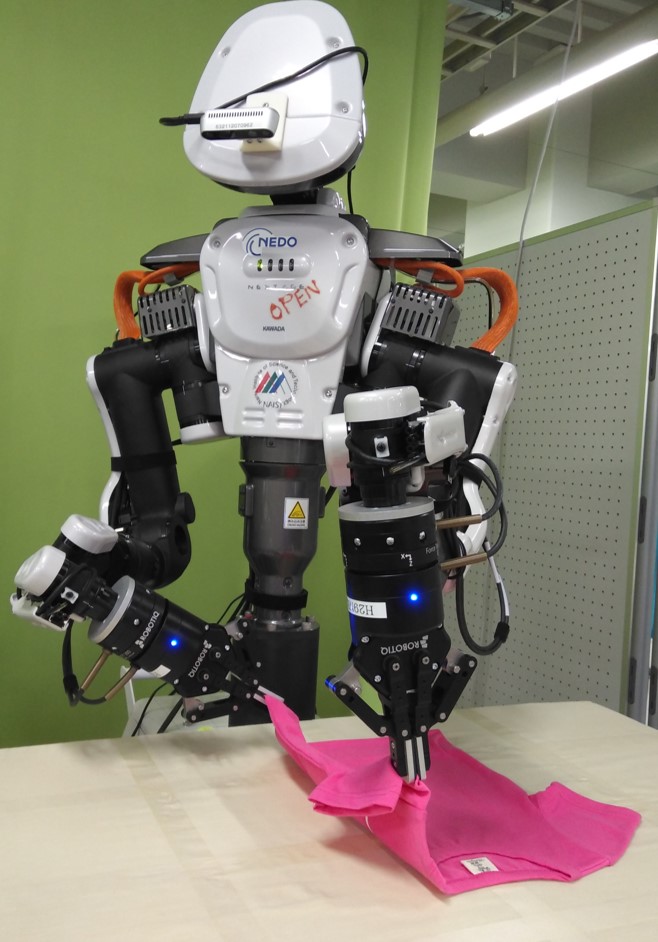}\label{fig_overview:fig1}}
\end{minipage}\hfill
\begin{minipage}{0.5\textwidth}
\centering
\subfloat[Handkerchief]{\includegraphics[width=4.1cm]{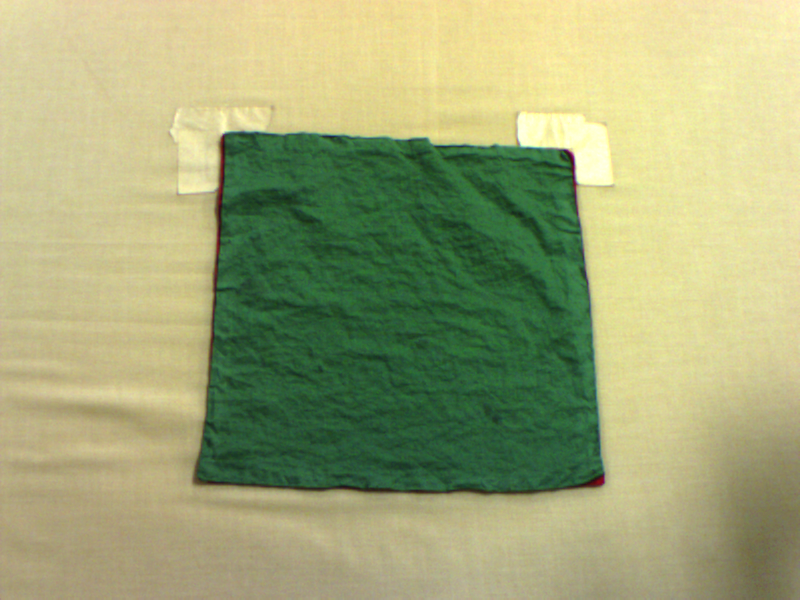}\label{fig_overview:fig2}}

\subfloat[Shirt]{\includegraphics[width=4.1cm]{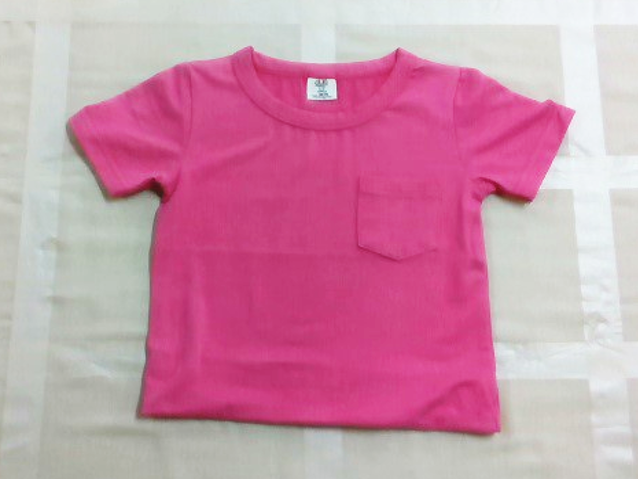}\label{fig_overview:fig3}}

\subfloat[Shorts]{\includegraphics[width=4.1cm]{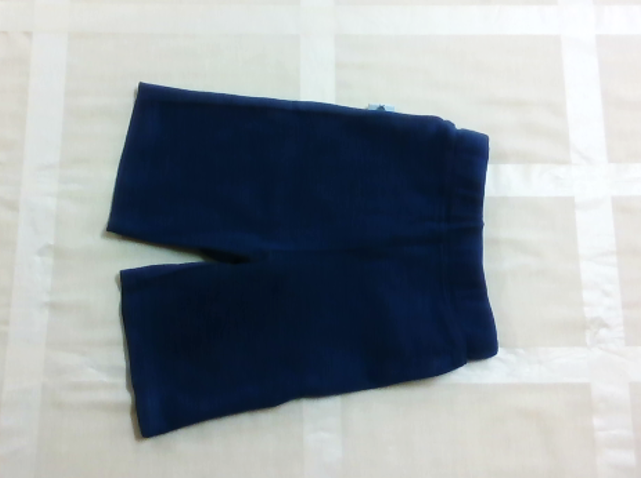}\label{fig_overview:fig4}}
\end{minipage}
\caption[]{Real-robot setting: Our targets are two robotic cloth-manipulation tasks with a dual-arm humanoid robot called NEXTAGE: (a) 1) turning a handkerchief over (b) and 2) folding a shirt (c) and a pair of shorts (d).}\label{fig_overview}
\end{figure}

We applied GA-GAIL to two goal-reaching tasks to evaluate learning performances and compared the effect of different demonstrations, discriminator settings, and goal label section methods. GA-GAIL was applied to robotic cloth-manipulation tasks to control a dual-arm humanoid robot called NEXTAGE (Fig. \ref{fig_overview:fig1}) to learn (1) to turn over a handkerchief (Fig. \ref{fig_overview:fig2}) and (2) to fold a shirt (Fig. \ref{fig_overview:fig3}) and a pair of shorts (Fig. \ref{fig_overview:fig4}). We chose a robotic cloth-manipulation task because it requires high-dimensional state definition and complicated reward function design. In the cloth-manipulation task, GA-GAIL uses a designed discrete action set, such as grasp-release and folding lines, to learn the policies with high task success rates from human demonstrations, including decision errors.

The contributions of this paper are summarized as follows: 
\begin{enumerate}
\item proposed a novel GAIL framework and added a second discriminator, the goal discriminator;

\item investigated learning performances from imperfect demonstrations in simulations and real-robot experiments;

\item investigated the performances of different combinations of each discriminator in simulations;

\item conducted real-robot experiments, which folded a shirt and a pair of shorts with human demonstrations that included errors.
\end{enumerate}

The remainder of this paper is organized as follows. Section \ref{sec:rel} describes previous research on GAIL and a robot's manipulation of cloth. Preparations are introduced in Section \ref{sec:prepar}. Our proposed method's details are explained in Section \ref{sec:propose}. Sections \ref{sec:sim} and \ref{sec:real} present our experimental results in simulations and real-robot experiments on cloth-manipulation tasks. A discussion and a conclusion are described in Sections \ref{sec:dis} and \ref{sec:con}.

\section{Related work}\label{sec:rel}
\subsection{Robotic cloth manipulation}
Robotic cloth manipulation can be divided into two approaches: task-oriented and knowledge-based \cite{yamazaki2018emd}. We focus on the latter that learns various cloth manipulation from data. 
Previous studies on learning cloth manipulation from label and demonstration data proposed learning of clothing type discrimination \cite{DBLP:conf/icra/KampourisMPSKTM16, Classification3DPointClouds18}, the manipulation of grasping points \cite{doumanoglou2016folding, DBLP:journals/pr/CoronaAGT18}, and cloth-manipulation policies \cite{yang2017repeatable,BedMaking19}.
Other methods learn a clothing transition model from manipulation data collected on actual environments and simulations \cite{yamazaki2018emd, yamazaki2019emd}. Lee et al. \cite{lee2015learning} presented force-based demonstration learning for deformable object manipulation. Matsubara et al. \cite{matsubara2013reinforcement} proposed a reinforcement learning (RL) approach to learn motor skills to handle a t-shirt based on the topological relationship between the robot configuration and non-rigid materials. Another friction-model-based RL approach searched for robot motion trajectories to wrap a scarf around a mannequin's neck \cite{FrictionModelBasedRL}. Jangir et al. \cite{DynamicClothDRL} proposed a method for learning a dynamic cloth-manipulation policy by observing a hem's coordinates and velocity in a simulation environment. Matas et al. \cite{pmlr-v87-matas18a} successfully applied a learning policy in simulations to a real-world cloth manipulation. 

We compare our propose method with the previous methods. It can learn high-performance policies from imperfect demonstrations using labeled goal states better than the cloth manipulation using imitation learning \cite{yang2017repeatable, BedMaking19,lee2015learning}. On the other hand, compared to the previous methods using RL \cite{RASDPN,matsubara2013reinforcement,FrictionModelBasedRL,DynamicClothDRL}, our method does not require the design of a complex reward function for each cloth-manipulation task.

\subsection{Generative adversarial imitation learning}
Generative Adversarial Imitation Learning (GAIL) \cite{GAIL} imitates demonstration policies by the adversarial learning of a generator and a discriminator. Previous GAIL studies proposed methods that improve sample efficiency \cite{NIPS2018_7972,DBLP:conf/aistats/BlondeK19,schroecker2018generative} and incorporated the structure of inverse reinforcement learning into a discriminator \cite{fu2018learning,qureshi2018adversarial}. Peng et al. \cite{peng2018variational} presented a stable GAIL framework that adjusts the discriminator's learning rate. Many of these methods update policies with a policy-search-based DRL. On the other hand, Reddy et al. \cite{SQIL} proposed a value-based GAIL framework that updates a policy with a value-based DRL. Compared to other GAIL methods, our proposed method learns control policies from imperfect demonstrations and goal state labels collected at low cost. 

Moreover, enhancing the discriminator for further performance improvement of GAIL has been studied. 
Kinose et al. \cite{AkiraKinose2020} integrally represented reinforcement learning and imitation learning with probabilistic graphical models and proposed a method for learning policies from imitation rewards and task-dependent achievement rewards. Ding et al. \cite{NIPS2019_9667} accelerated the convergence of policies by updating the goal state in demonstration samples based on the learning progress. Sun et al. \cite{DBLP:conf/ijcai/SunM19} proposed a GAIL framework from demonstrations with imperfect action sequences. This framework learns imitation policies for state trajectories while using demonstration actions as auxiliary information. Another approach to learning from imperfect demonstrations is to optimize demonstration weights without labeling information\cite{DBLP:conf/icml/WuCBTS19, pmlr-v130-tangkaratt21a}. A feature of our proposed method is the inclusion of goal information in addition to demonstrations. This strategy improves the task success rate of imitation policies with a goal discriminator in a reaching task.

\section{Preparation}\label{sec:prepar}
\subsection{Reinforcement learning}
Reinforcement learning (RL) \cite{sutton1998reinforcement,kober2013reinforcement} solves the Markov decision process (MDP) defined by a 5-tuple $(\mathcal{S}, \mathcal{A}, \mathcal{T}, \mathcal{R}, \gamma)$. $\mathcal{S} = \{s_1, s_2, ..., s_n\}$ is a finite set of states. $\mathcal{A}= \{a_1, a_2, ..., a_m\}$ is a finite set of actions. $\mathcal{T}_{s s'}^{a}$ is the probability of transitioning from state $s$ to state $s'$ under action $a$. The corresponding reward is defined as $r_{s s'}^{a}$ with reward function $\mathcal{R}$. $\gamma \in (0, 1)$ is a discount parameter. Policy $\pi(a|s)$ represents the probability of action $a$ being taken under state $s$. The value function is defined as the expected discounted total reward in state $s$:
\begin{eqnarray}
\begin{split}
V(s)={\mathbb{E}}_{\pi, \mathcal{T}}\bigg[\sum_{\substack{t = 0}}^{\infty} \gamma^{t} r_{s_{t}} \bigg| s_{0}= s \bigg],
\label{V_function}
\end{split}
\end{eqnarray}
where $r_{s_{t}} = \sum_{\substack{a \in \mathcal{A} \\ s' \in \mathcal{S}}}\pi(a|s_{t}) \mathcal{T}_{s_{t}s'}^{a} r_{s_{t}s'}^{a}$ is the expected reward from state $s_{t}$.

RL's objective is to find optimal policy $\pi^{*}$ that maximizes the value function to satisfy the following Bellman equation:
\begin{eqnarray}
\begin{split}
V^{*}(s) = \displaystyle\max_{\pi} \sum_{\substack{a \in \mathcal{A} \\ s' \in \mathcal{S}}} \pi(a|s) \mathcal{T}_{ss'}^{a} \big(r_{ss'}^{a} + \gamma V^{*}(s')\big),
\label{V_Bellman}
\end{split}
\end{eqnarray}
or a Q function for state-action pairs ($s, a$):
\begin{eqnarray}
\begin{split}
Q^{*}(s, a) = \displaystyle\max_{\pi} \sum_{s' \in \mathcal{S}}\mathcal{T}_{ss'}^{a}\big(r_{ss'}^{a}+  \gamma  \sum_{a' \in \mathcal{A}}  \pi(a'|s')Q^{*}(s', a')\big).
\label{Q}
\end{split}
\end{eqnarray}

Value-based RL algorithms, e.g., Q-learning~\cite{watkins1992q}, SARSA~\cite{sutton1996generalization}, and LSPI~\cite{lagoudakis2003least}, approximate the value/Q function using Temporal Difference (TD) error. For example, the TD update rule in Q-learning follows $Q(s_t, a_t) \gets Q(s_t, a_t) + \alpha[r_{s_{t}s_{t+1}}^{a_{t}}+ \gamma \max_{a_{t+1}}Q(s_{t+1}, a_{t+1})-Q(s_t, a_t)]$, where $\alpha$ is the learning rate.

\subsection{Dynamic policy programming}\label{pr_DPP}
To exploit the nature of smooth policy updates, Dynamic Policy Programming (DPP) \cite{azar2011dynamic,azar2012dynamic} considers the Kullback-Leibler divergence between current policy $\pi$ and baseline policy $\bar{\pi}$ in the reward function to minimize the difference between the current and baseline policies while maximizing the expected reward:
\begin{eqnarray}
D_{\mathrm{KL}} = \sum_{\substack{a \in \mathcal{A}}} \pi (a|s) \log \frac{\pi(a|s)}{\bar{\pi}(a|s)}.
\label{}
\end{eqnarray}
Thus, the Bellman optimality equation in Eq. (\ref{V_Bellman}) is modified:
\begin{eqnarray}
\begin{split}
V_{\bar{\pi}}^{*}(s)=\max_{\pi}\sum_{\substack{a \in \mathcal{A} \\s' \in \mathcal{S}}}\pi(a|s)\bigg[\mathcal{T}_{ss'}^{a}  \big(r_{ss'}^{a}+\gamma V^{*}_{\bar{\pi}}(s')\big) - \frac{1}{\eta}\log\Big(\frac{\pi(a|s)}{\bar{\pi}(a|s)}\Big)\bigg].
\label{New_Value_Opt}
\end{split}
\end{eqnarray}
The effect of the Kullback-Leibler divergence is controlled by inverse temperature $\eta$. Following a previous work \cite{azar2011dynamic,todorov2006linearly}, we let $\eta$ be a positive constant.
Optimal value function $V_{\bar{\pi}}^{*}(s)$ for all $s \in \mathcal{S}$ and optimal policy $\bar{\pi}^{*}(a|s)$ for all $(s, a)$ satisfy the following double-loop fixed-point iterations:
\begin{eqnarray}
\begin{split}
V_{\bar{\pi}}^{t+1}(s)=\frac{1}{\eta} \log \sum_{\substack{a \in \mathcal{A}}}  \bar{\pi}^{t} (a|s) \exp \Big[\eta   \sum_{\substack{s' \in \mathcal{S}}}  \mathcal{T}_{ss'}^{a}  \big(r_{ss'}^{a} + \gamma V_{\bar{\pi}}^{t}(s')\big) \Big]
\label{Close_Form_V}
\end{split}
\end{eqnarray}

\begin{eqnarray}
\begin{split}
\bar{\pi}^{t+1}(a|s) =  \frac{\bar{\pi}^{t}(a|s)\exp\Big[\eta\sum\limits_{\substack{s' \in \mathcal{S}}}\mathcal{T}_{ss'}^{a}\big(r_{ss'}^{a} + \gamma V_{\bar{\pi}}^{t}(s')\big)\Big]}{\exp\big(\eta V_{\bar{\pi}}^{t+1}(s)\big)}.
\label{Close_Form_Pi}
\end{split}
\end{eqnarray}

An action preference function \cite{sutton1998reinforcement} at the $(t+1)$-iteration for all state-action pairs ($s, a$) is defined following a previous work \cite{azar2012dynamic} to obtain the optimal policy that maximizes the above value function:
\begin{eqnarray}
\begin{split}
P_{t+1}(s, a) = \frac{1}{\eta}\log\bar{\pi}^{t}(a|s) + \sum_{s' \in \mathcal{S}} \mathcal{T}^{a}_{ss'}\big(r^{a}_{ss'}+\gamma V^{t}_{\bar{\pi}}(s')\big).
\label{Action_preference}
\end{split}
\end{eqnarray}
Combining Eq. (\ref{Action_preference}) with Eqs. (\ref{Close_Form_V}) and (\ref{Close_Form_Pi}) the following simple form is obtained:
\begin{eqnarray}
\begin{split}
V_{\bar{\pi}}^{t}(s)= \frac{1}{\eta}\log{\sum_{a^{} \in \mathcal{A}}}\exp\big(\eta P_{t}(s, a)\big)
\label{DPP_V}
\end{split}
\end{eqnarray}

\begin{eqnarray}
\begin{split}
\bar{\pi}_{}^{t}(a|s) = \frac{\exp\big(\eta P_{t}(s, a)\big)}{\sum_{a' \in \mathcal{A}}\exp\big(\eta P_{t}(s, a')\big)}.
\label{DPP_pi}
\end{split}
\end{eqnarray}
The update rule of action preference function $P_{t+1}(s, a)={\mathcal{O}}P_{t}(s, a)$ is derived by plugging Eqs. (\ref{DPP_V}) and (\ref{DPP_pi}) into Eq. (\ref{Action_preference}):
\begin{eqnarray}
\begin{split}
\mathcal{O} P_t (s, a) = P_{t}(s, a) -{\mathcal{L}}_{\eta} P_{t}(s) + \sum_{s' \in \mathcal{S}}\mathcal{T}^{a}_{ss'}\big(r^{a}_{ss'}+\gamma {\mathcal{L}}_{\eta} P_{t}(s')\big),
\label{DPP_recursion_L}
\end{split}
\end{eqnarray}
where ${\mathcal{L}}_{\eta} P_{}(s) \triangleq \frac{1}{\eta}\log{\sum_{a^{} \in \mathcal{A}}}\exp(\eta P_{}(s, a)) = V_{\bar{\pi}}(s)$.
The difference between $P_{t+1}(s, a)$ and $\mathcal{O} P_t(s, a)$ is used to calculate the error signal to train the action preference function.

\subsection{Generative adversarial imitation learning}
The Generative Adversarial Imitation Learning (GAIL) framework consists of two parts: a generator that learns the sampling distribution from a demonstrator and a demonstration discriminator that distinguishes between generated and demonstration samples. Under the adversarial framework, a sampling distribution is ideally learned that is indistinguishable to a demonstration, i.e., the demonstration discriminator learns classification ability while the generator learns to confuse the demonstration discriminator. 

Given these ideas, GAIL's objective is formulated:
\begin{equation}
  \max_{\pi} \min_{D} \mathbb { E } _ { \pi } [ - \log ( 1 -  D_D( s , a ) ) ] + \mathbb { E } _ { \pi _ { D } } [ - \log ( D_D ( s , a ) ) ]. 
\label{ep:gail}
\end{equation}
Given current state $s$ and action $a$ as input, the discriminator outputs the probability that the input belongs to demonstration $D_D( s , a ) \in [0, 1]$. $\pi$ is a learning policy for the generator, and $\pi_{D}$ is a demonstration policy. Since a Deep Neural Network (DNN) approximates the generator and the demonstration discriminator, we take a gradient-based numerical simultaneous optimization approach.  

The discriminator's loss function is defined:
\begin{equation}
  \mathbb { E } _ { \pi } [ - \log ( 1 -  D_D( s , a;\phi ) ) ] + \mathbb { E } _ { \pi _ { D } } [ - \log ( D_D( s , a;\phi ) ) ].
\label{ep:d_loss}
\end{equation}
where $\phi$ is defined as the DNN parameters that approximate the demonstration discriminator. Since the demonstration discriminator is a simpler function than the generator's policy, its learning progress is faster than the policy. Hence, parameters $\phi$ must be updated conservatively to avoid over-fitting.

Given current demonstration discriminator $D_D(s,a;\phi)$, updated policy $\pi$ of the generator is formulated:
\begin{equation}
  \pi^\ast (a|s) = \argmax_{\pi} \mathbb { E } _ { \pi } [ - \log ( 1 -  D_D( s , a;\phi ) ) ].
\label{ep:gail_pi_opt}
\end{equation}
Using $r_{s}^a = - \log ( 1 - D_D( s , a; \phi) )$ in the reward function, the RL agent learns a policy that maximizes the total reward. 

Most previously proposed GAIL frameworks employ Trust Region Policy Optimization (TRPO)~\cite{pmlr-v37-schulman15} in the generator, a popular policy search based RL approach. 
It is highly suitable for GAIL due to two key features: 1) a smooth policy update for learning stability and 2) the policy's diversity for sampling a wide range for training the discriminator. 
In a benchmark of discrete action spaces \cite{SonicBenchmark}, a value-based DRL outperformed Proximal Policy Optimization \cite{DBLP:journals/corr/SchulmanWDRK17}, which is a variant of TRPO. Due to this reason, existing GAIL frameworks may be unsuitable for our purpose, i.e., for robotic cloth manipulation with a discrete action set. Although value-based DRLs have been proposed \cite{haarnoja2017reinforcement,RASDPN}, they might also be inappropriate because they do not simultaneously have the two key features. 
Therefore, our proposed GAIL framework employs a modified value-based DRL, the Entropy-maximizing Deep P-Network (EDPN), which can consider both smoothness and causal entropy in policy updates.

\section{Proposed method}\label{sec:propose}

\begin{figure}
  \begin{center}
    \includegraphics[clip,scale=0.6]{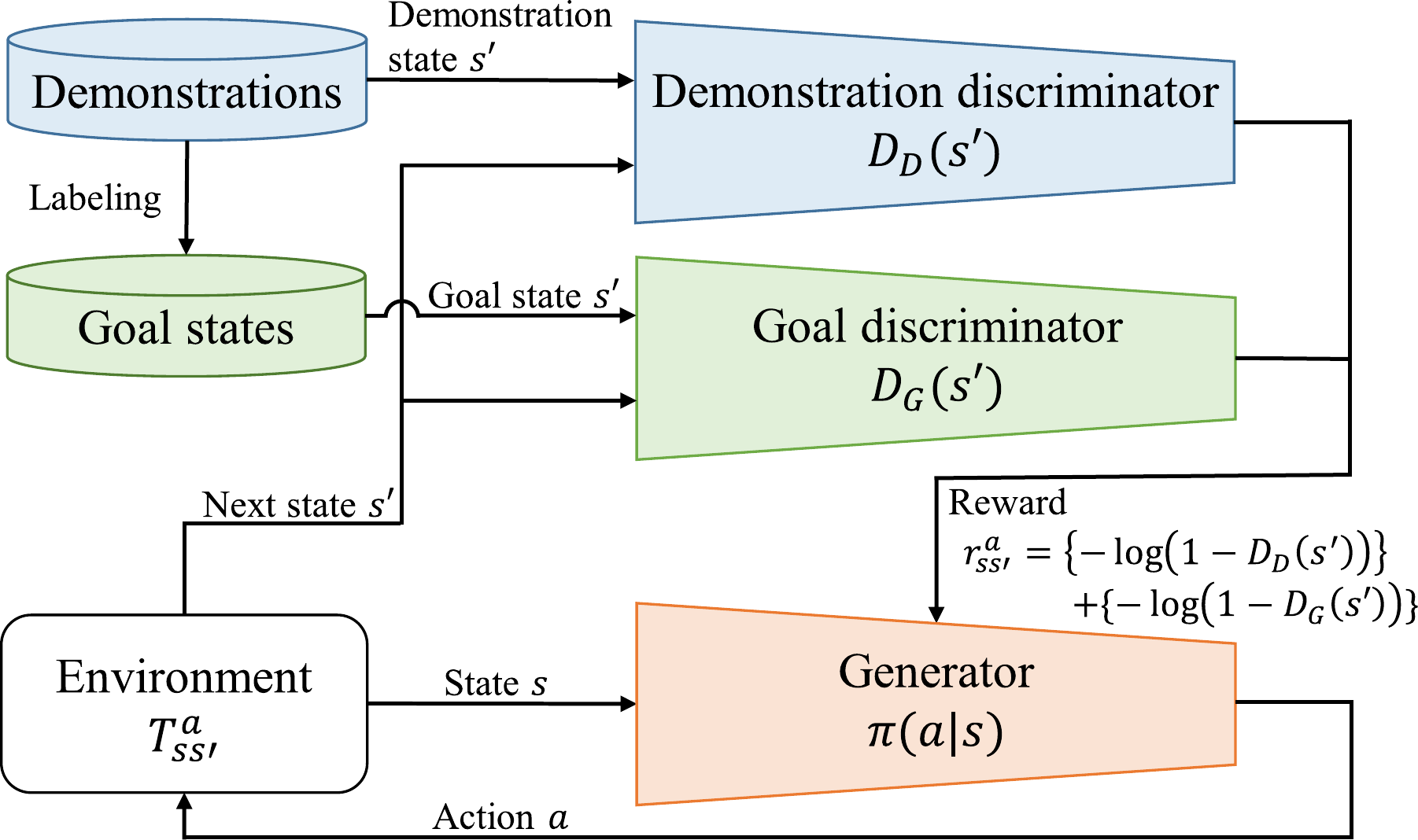}
    \caption{Overview of proposed GA-GAIL}
    \label{fig:overview_ddGAIL}
  \end{center}
\end{figure}

\subsection{GA-GAIL overview}
As shown in Fig. \ref{fig:overview_ddGAIL}, GA-GAIL consists of the following three components:
\begin{itemize}
\item demonstration discriminator $D_D(s)$, which distinguishes between demonstration and generated states;
\item goal discriminator $D_G(s)$, which distinguishes between goal and generated states;
\item generator $\pi(a|s)$, which learns policies to achieve the goal state while imitating the demonstration.
\end{itemize}
Comparing GA-GAIL and the standard GAIL framework, we added a goal discriminator to GA-GAIL whose reward function consists of demonstration and goal discriminators. Since goal-sample $\mathcal{M}_G$ is a state selected from within demonstrations $\mathcal{M}_D$, the relation is $\mathcal{M}_G \subset \mathcal{M}_D$. Two discriminators learn classification ability while the generator learns to confuse the two discriminators. 

A previous GAIL study \cite{DBLP:conf/ijcai/SunM19} reduced the negative effects of imperfect demonstrations by minimizing the state distribution of the demonstration and learning policies while using demonstration actions as an auxiliary. These previous studies successfully learned imitation policies by inputting only the state into the discriminator \cite{DBLP:conf/ijcai/SunM19,DBLP:journals/corr/MerelTTSLWWH17}. In this study, we follow those studies and only use the state as input to demonstration discriminator $D_D(s)$ to reduce the negative effects of imperfect demonstrations. In our simulation experiments (Section \ref{subsec:disc_set}), the proposed method's performance is also improved when a demonstration discriminator is not inputted with the action.

GA-GAIL's objective is formulated:
\begin{equation}
\begin{split}
  \max_{\pi} \min_{D_D} \min_{D_G} \quad &\mathbb { E } _ { \pi } [ - \log ( 1 -  D_D( s ) ) - \log ( 1 -  D_G( s ) ) ]\\ &+ \mathbb { E } _ { \pi _ { D } } [ - \log ( D_D( s ) ) ] + \mathbb { E } _ { d_G } [ - \log ( D_G( s ) ) ]. 
\label{ep:gail}
\end{split}
\end{equation}
Given state $s$ as input, two discriminators output probability $D_D(s) \in [0, 1], D_G(s) \in [0, 1]$. $\pi$ is the learning policy for the generator, and $\pi_{D}$ is the demonstration policy. $d_G(s)$ is the stationary distribution of the observed goal state. Since the generator and both discriminators are approximated by DNN, we take a gradient-based numerical simultaneous optimization approach, which is identical as that for GAIL.  
More details of the optimization are given in subsequent sections.

\subsection{Discriminator optimization}
Next we define loss function of the two discriminators:
\begin{equation}
  J_D(\phi) = \mathbb { E } _ { \pi } [ - \log ( 1 -  D_D( s ; \phi) ) ] + \mathbb { E } _ { \pi _ { D } } [ - \log ( D_D( s ; \phi) ) ]
\label{ep:d_e_loss}
\end{equation}
\begin{equation}
  J_G(\omega) = \mathbb { E } _ { \pi } [ - \log ( 1 -  D_G( s ; \omega) ) ] + \mathbb { E } _ { d_G } [ - \log ( D_G( s ; \omega) ) ].
\label{ep:d_t_loss}
\end{equation}
where $\phi, \omega$ is defined as the DNN parameters that approximate the discriminator. 
Parameters $\phi, \omega$ are updated with the stochastic gradient method. Since the discriminator can be a simpler function than the policy, the learning progress is faster than the policy. 
Thus, parameters $\phi, \omega$ must be updated conservatively to avoid over-fitting. 
The parameters and network structure used in the our study's experiments are shown in Section \ref{subsec:disc_para}.

\subsection{Generator optimization}
Given demonstration discriminator $D_D(s)$ and goal discriminator $D_G(s)$, updated policy $\pi$ of the generator is formulated:
\begin{equation}
  \pi^\ast = \argmax_{\pi} \mathbb { E } _ { \pi } [ - \log ( 1 -  D_D( s ) ) - \log ( 1 -  D_G( s ) ) ]. 
\label{ep:gail_pi_opt}
\end{equation}
RL's reward function is formulated:
\begin{equation}
  r_{s'} = \{ - \log ( 1 -  D_D( s' ) ) \} + \{  - \log ( 1 -  D_G( s' ) ) \}.
\label{ep:reaching_reward}
\end{equation}
The reward is calculated from next state $s'$ and the two discriminators. 
To stably learn the policy from the reward by changing the learning of both discriminators, the policy must be smoothly updated while maintaining its diversity. GA-GAIL learns the generator's policy by EDPN by following DPN with an important modification so that it has two such properties as smooth update and diversity in its policy. To this end, the reward function is designed:
\begin{equation}
  r_{EDPN} = r_{s'} - \frac{1}{\eta} {\mathrm{KL}}(~\pi \parallel \bar {\pi}~) + \sigma {H}(~\pi~),
  \label{ep:EDPN_reward}
\end{equation}
where 
\begin{eqnarray}
  {\mathrm{KL}}(~\pi \parallel \bar {\pi}~) = \sum_{\substack{a \in \mathcal{A}}} \pi (a|s) \log \frac{\pi(a|s)}{\bar{\pi}(a|s)}
  \label{ep:KL}
\end{eqnarray}
and 
\begin{equation}
  {H}(~\pi~) = \sum _{a\in \mathcal{A}} -\pi(a|s) \log \left(\pi(a|s) \right). 
   \label{ep:entropy}
\end{equation}
The second term promotes a smooth policy update, whose amount is quantified by Kullback-Leibler divergence ${\mathrm{KL}}(\pi \parallel \bar {\pi})$ between current policy $\pi$ and baseline policy $\bar{\pi}$. The third term promotes the diversity of the actions in the policy and calculates entropy ${H}(\pi)$. $\eta$ and $\sigma$ are coefficients that control the balance between Kullback-Leibler divergence ${\mathrm{KL}}(\pi \parallel \bar {\pi})$ and entropy ${H}(\pi)$. By learning a policy that maximizes this total reward, the policy is smoothly updated while maximizing its entropy.
Value-based reinforcement learning with these two constraints has been robust to the approximation error of value functions \cite{pmlr-v89-kozuno19a}. 

Following a previous study \cite{azar2011dynamic}, we derived an update rule of EDPN's action preferences function is derived: 
\begin{equation}
    P_{t+1} (s,a) = \frac{1}{1+\sigma \eta}(P_t(s,a) - V_{\bar \pi}^{t} (s)) + 
    \sum_{s\in \mathcal{S}} T_{ss'}^a \left(r_{s'} + \gamma V_{\bar \pi}^{t} (s') \right),
    \label{ep:EDPN_def_next_p}
\end{equation}
where
\begin{equation}
    \bar \pi_t (a|s) = 
    \frac {\exp ( \frac{\eta}{1+\sigma \eta} P_t(s,a)) } {\sum _{a'\in \mathcal{A}}\exp ( \frac{\eta}{1+\sigma \eta} P_t(s,a') ) }
    \label{ep:EDPN_new_policy}
\end{equation}
and
\begin{equation}
    V_{\bar \pi}^{t} (s) = \frac{1+\sigma \eta}{\eta} \log \sum _{a\in \mathcal{A}}\exp ( \frac{\eta}{1+\sigma \eta} P_t(s,a) ).
    \label{ep:EDPN_new_v}
\end{equation}
$T_{ss'}^a$ is the probability of transitioning from state $s$ to next state $s'$ under action $a$. $\gamma \in (0, 1)$ is a discount parameter. 

When learning a policy for image-based inputs, the action preferences function is approximated by DNN. $\theta$ is defined as the DNN parameters that approximate the action preference function and calculate teaching signal $y(\theta^{-})$ following Eq. (\ref{ep:EDPN_def_next_p}):
\begin{equation}
  y(\theta^{-}) = \frac{1}{1+\sigma \eta}(\hat{P_t}(s,a;\theta^{-}) - \hat{V}_{\bar \pi}^{t} (s;\theta^{-})) + 
   r_{s'} + \gamma \hat{V}_{\bar \pi}^{t} (s';\theta^{-}).
  \label{ep:EDPN_target}
\end{equation}

The current parameters as $\theta^{-}$ are saved to build a target network. Then the gradient of following loss function $J_{}(\theta, \theta^-)$ is computed and used to update parameter $\theta$ using the stochastic gradient descent: 
\begin{equation}
J_{}(\theta, \theta^-)\triangleq(y(\theta^-)-\hat{P}(s, a; \theta))^2.
\label{ep:EDPN_Loss}
\end{equation}

Note that if $\sigma=0$, EDPN becomes equivalent to Deep P-Network (DPN) \cite{RASDPN}, then EDPN is a generalized version of DPN that is suitable for GAIL.

\subsection{Summary}
The pseudocode for training GA-GAIL is shown in Algorithm \ref{alg:imitation}. 
The training of GA-GAIL has four steps in one iteration, which are executed a fixed number of times: 
\begin{enumerate}
  \item generate new samples following current policy $\pi$ based on current action preference function $P(s,a)$;
  \item update two discriminators by updating $\phi, \omega$ to minimize Eqs. (\ref{ep:d_e_loss}) and (\ref{ep:d_t_loss});
  \item calculate the reward for samples stored in local memory $\mathcal{M}$ following both updated discriminators $D_D, D_G$;
  \item update current policy $\pi$ using EDPN.
\end{enumerate}

\begin{algorithm}
    \SetKwData{Left}{left}\SetKwData{This}{this}\SetKwData{Up}{up}
  \SetKwFunction{Union}{Union}\SetKwFunction{FindCompress}{FindCompress}
  \SetKwInOut{Input}{input}\SetKwInOut{Output}{output}
  
  Initialize GA-GAIL iteration number $I$, generator parameters $M,T$, network iteration number $J,K$, local memory $\mathcal{M}$ and its size $E$\\
  Initialize number of minibatches $B = E/(minibatch\_size)$\\
  Load demonstrations $\mathcal{M}_D$ and goal labels $\mathcal{M}_G$\\
  Initialize value network weights $\theta$, target network weights $\theta^- = \theta$, discriminator network weights $\phi, \omega$\\
  \For{$ i = 1, 2, ..., I$}
  {
  \tcp{Generate new samples with current policy $\pi_i$}
  \For{$episode = 1, 2, ..., M$}
  {
      \For{$t = 1, 2, ..., T$}
      {
        Take action $a$ with softmax policy based on $\pi_i$ following $P^i(s,a;\theta)$ and Eq. (24)\\
        Receive new state $s'$\\
        Store transition $(s, a, s')$ in $\mathcal{M}$
      }
  }
  \tcp{Update two discriminators}
  \For{$j = 1, 2, ..., J$}
  {
    Shuffle local memory $\mathcal{M}$ index\\
    \For{$b = 1, 2, ..., B$}
    {
        Sample minibatch of generator's state $\tau$ in $\mathcal{M}$\\
        Sample minibatch of demonstration's state $\tau_D$ in $\mathcal{M}_D$\\
        Sample minibatch of goal state $\tau_G$ in $\mathcal{M}_G$\\
        Get loss of demonstration discriminator: $J_D(\phi; \tau, \tau_D) = \mathbb { E } _ { \tau } [ - \log ( 1 -  D^i_D( s ; \phi) ) ] + \mathbb { E } _ { \tau_D } [ - \log ( D^i_D( s ; \phi) ) ]$\\
        Get loss of goal discriminator: $J_G(\omega; \tau, \tau_G) = \mathbb { E } _ { \tau } [ - \log ( 1 -  D^i_G( s ; \omega) ) ] + \mathbb { E } _ { \tau_G } [ - \log ( D^i_G( s ; \omega) ) ]$\\
        Update $\phi, \omega$ by performing a gradient descent step on $J_D(\phi; \tau, \tau_D)+J_G(\omega; \tau, \tau_G)$
    }
  }
  \tcp{Calculate the reward for samples}
  \For{$b = 1, 2, ..., B$}
  {
    Sample minibatch of transition $(s, a, s')$ in $\mathcal{M}$\\
    Calculate minibatch of reward: $r_{s'} = \{ - \log ( 1 -  D^i_D( s'; \phi ) ) \} + \{  - \log ( 1 -  D^i_G( s'; \omega ) ) \}$\\
    Store transition $(s, a, s', r_{s'})$ in $\mathcal{M}$
  }
  \tcp{Update current policy $\pi_i$ using EDPN}
  \For{$k = 1, 2, ..., K$}
  {
    Shuffle local memory $\mathcal{M}$ index\\
    \For{$b = 1, 2, ..., B$}
    {
      Sample minibatch of transition $(s, a, s', r_{s'})$ in $\mathcal{M}$\\
      Calculate the teaching signal: $y_b = \frac{1}{1+\sigma \eta}(\hat{P_i}(s,a;\theta^{-}) - \hat{V}_{\bar \pi}^{i} (s;\theta^{-})) + r_{s'} + \gamma \hat{V}_{\bar \pi}^{i} (s';\theta^{-})$ \\
      Get loss and update $\theta$ by performing a gradient descent step on $(y_b-\hat{P}(s,a;\theta))^2$
    }
  }
  Update target network $\theta^- = \theta$
  }
\caption{GA-GAIL}
\label{alg:imitation}
\end{algorithm}

\section{Simulation}\label{sec:sim}
\begin{figure}[t]
    \centering
    \subfloat[Overview of $2$-DoF manipulator twice-reaching task\label{fig:2dof_twice}] {\includegraphics[width=0.7\columnwidth]{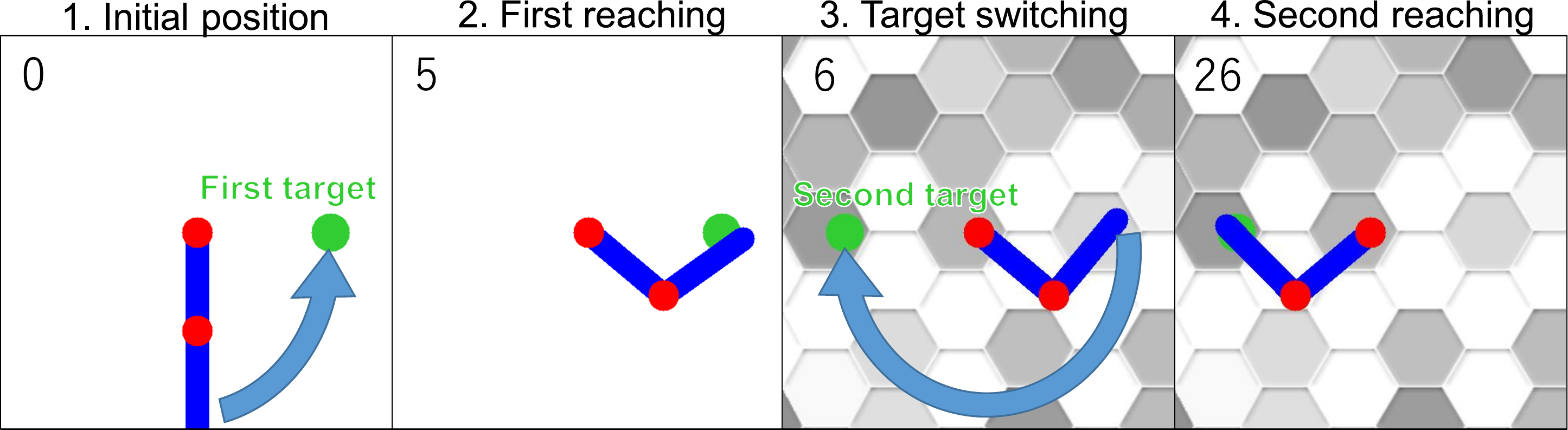}}\\
    \subfloat[Overview of pick-and-place task\label{fig:pick_and_place}]{\includegraphics[width=0.7\columnwidth]{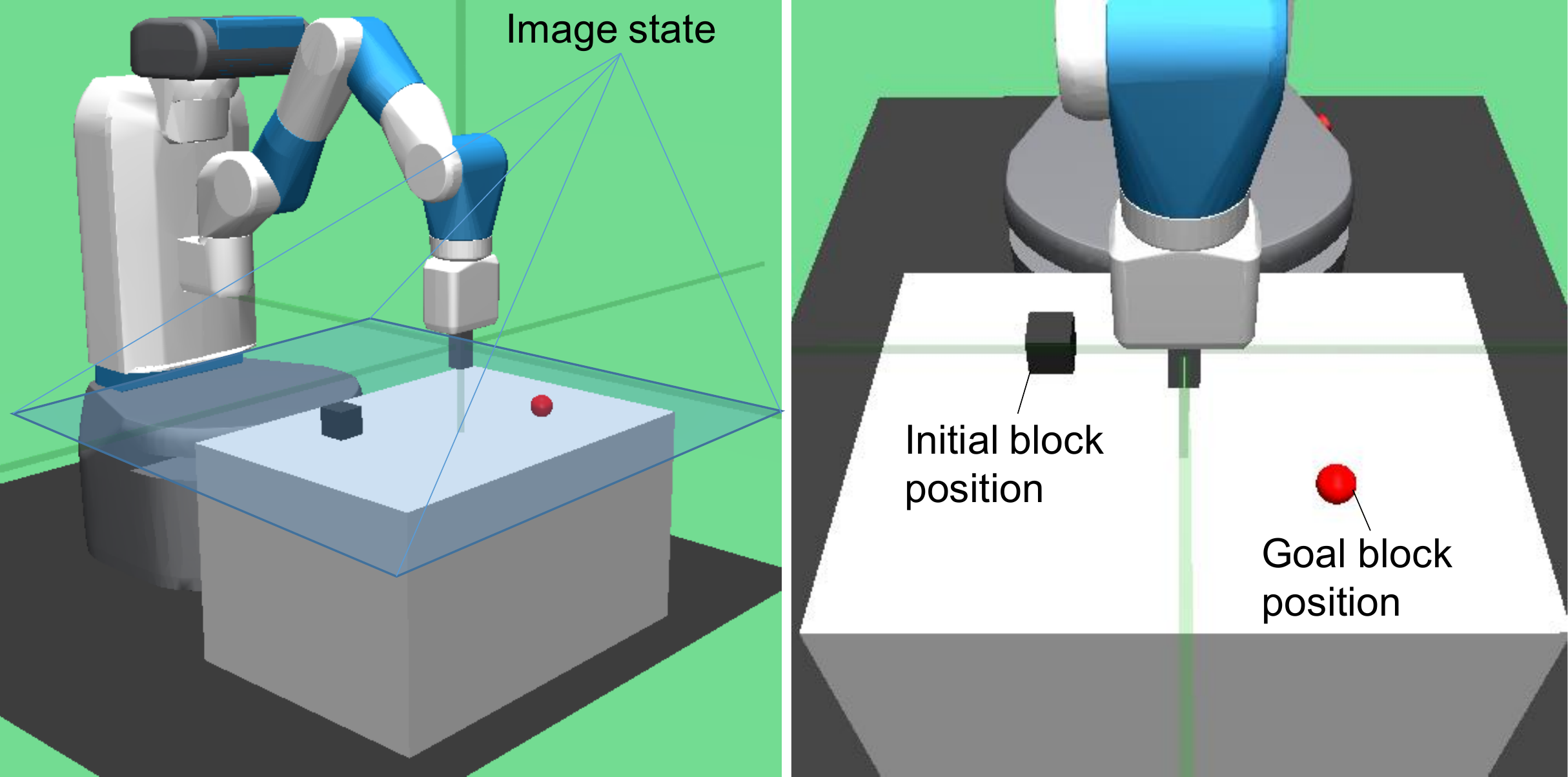}}
    \caption{Two baseline tasks used in simulation verification: In $2$-DoF manipulator twice-reaching task (a), background image and target position change when manipulator tip is close to first target. Number in upper left corner indicates steps. In pick-and-place task (b), a robot moves a block to goal position using image as an observation.}
    \label{fig:sim_tasks}
\end{figure}

In this section, we clarified the following two questions through two simulation tasks shown in Fig. \ref{fig:sim_tasks}:
\begin{enumerate}
\renewcommand{\labelenumi}{\arabic{enumi})}
\item Can our proposed method learn a high-performance policy from imperfect demonstrations?
\item Does combining two discriminators improve the learning policy's performance?
\end{enumerate}
For verifying question 1), we collected imperfect demonstrations from a DRL policy with a different number of updates. The number of updates to the DRL policy corresponds to the demonstration's imperfection. Question 1) is clarified by comparing the proposed method with other methods learned from demonstrations with different imperfections in simulations. We verified question 2) by comparing the performances of GA-GAIL, GA-GAIL without demonstration discriminator $D_D(s)$, and GA-GAIL without goal discriminator $D_G(s)$. Other simulation experiments compared our proposed method with RL from an imperfect demonstration, which uses a task-dependent reward function. GA-GAIL uses goal label information in addition to demonstrations, and we investigated the effect of the goal label selection method on learning policies. We applied the proposed method to two different reaching tasks to investigate its performance (Fig. \ref{fig:sim_tasks}).

\subsection{Settings}\label{sec:sim_setting}
This section describes the GA-GAIL settings as well as two baseline task settings in the simulation. The GA-GAIL discriminator settings are common to all the experiments, were described in the first subsection. Other settings for GA-GAIL are described in the subsections for each task.

\subsubsection{Discriminator settings}\label{subsec:disc_para}
\begin{figure}[t]
    \centering
    \includegraphics[width=0.8\columnwidth]{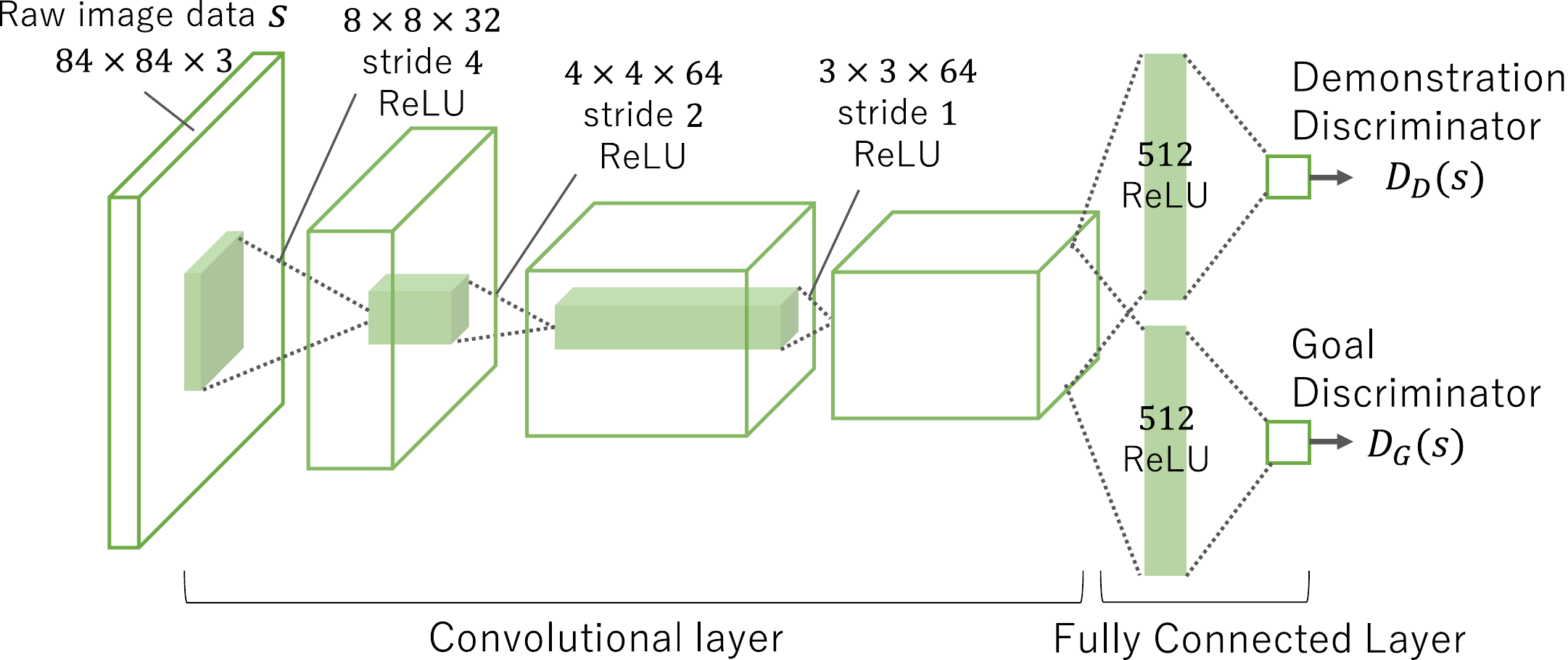}
    \caption{Network structure of demonstration discriminator $D_D(s)$ and goal discriminator $D_G(s)$: Input image is a $84 \times 84$ color image. Convolutional layer has three layers, and parameter $W \times W \times U$ for each layer, where $W$ is filter size and $U$ is the number of filters. Number shown in fully connected layer is the number of nodes.}
    \label{fig:disc_struct}
\end{figure}

The discriminator parameters, such as network structure and learning rate, are identical for all the experiments in this study. The network structure of demonstration discriminator $D_D(s)$ and goal discriminator $D_G(s)$ is shown in Fig. \ref{fig:disc_struct}. We employed RMSProp to optimize the parameters of the discriminator. The parameters of this optimizer are shown in Table \ref{table:disc_para}. 

\begin{table}[t]
  \centering
  \begin{tabular}{@{}p{3cm}p{1.25cm}l@{}}
      \toprule
      \textbf{Parameter} & \textbf{Value}  \\ \midrule
      Learning rate & $0.00025$ \vspace{0.2cm} \\
      Momentum & $0.95$ \vspace{0.2cm} \\
      Epsilon & $0.01$ \vspace{0.2cm} \\
      Clipping gradients & $5$ \vspace{0.2cm} \\
      Batch size & $32$\\
      \bottomrule
  \end{tabular}
   \caption{Parameters of RMSProp to optimize demonstration discriminator $D_D(s)$ and goal discriminator $D_G(s)$}
   \label{table:disc_para}
\end{table}

\subsubsection{$2$-DoF manipulator twice-reaching task}\label{sec:twice_tasks}
Figure \ref{fig:2dof_twice} shows an overview of the $2$-DoF manipulator twice-reaching task (twice-reaching task). This reaching task resembles uncovering the true target position by turning on a light switch's power at the first reaching, an action that switches on the background and target position. The environment and the GA-GAIL parameter settings are shown in Tables \ref{table:2doftwice_setting} and \ref{table:2doftwice_GAIL_setting}.

\begin{table}
    \begin{center}
        \subfloat[Parameter setting of $2$-DoF manipulator twice-reaching task\label{table:2doftwice_setting}]{
          \begin{tabular}{@{}p{3cm}p{8cm}l@{}}
          \toprule
          \textbf{MDP setting} & \textbf{Description}  \\ \midrule
          State & Input state is a $84 \times 84$ px color image. Number of state dimensions is $84 \times 84 \times 3 \,(21,168)$ \vspace{0.25cm} \\
          Action & Discrete actions $[-0.0875, -0.0175,$ $ 0, 0.0175, 0.0875]$ (rad) to increment joint with respective angle for each DoF. We define an action at each time step as one move per joint to reduce actions to $(n \times 5)$. \vspace{0.25cm} \\
          Reward for evaluation & Reward function is set as $r = -\big(|X_{\text{target}}-X|+|Y_{\text{target}}-Y|\big)$ where $X, Y$ is current position of the manipulator's end-effector, and $X_{\text{target}} = 0.6830, Y_{\text{target}} = 0$ is the target position. When distance between target and manipulator's tip is smaller than $0.2$, $5$ is added to the reward. \vspace{0.25cm} \\
          Initial state & First joint is set to position $[0, 0]$. All angles are initialized to $0$ rad simulation's start. \vspace{0.25cm} \\
          Demonstrations & Demonstrations are sampled from an RL policy that maximizes the reward for evaluation. Demonstrations ($30$ steps $\times$ $5$ trajectory = $150$ samples) are generated from exploring policies during learning process.\vspace{0.25cm} \\
          Goal state & Goal samples are selected from demonstrations, where distance between manipulator's end-effector and target position is less than $0.2$. \\
          \bottomrule
          \end{tabular}
        }

        \subfloat[Parameter setting of GA-GAIL algorithm\label{table:2doftwice_GAIL_setting}]{
          \begin{tabular}{@{}lp{8cm}llll@{}}
          \toprule
           \textbf{Parameter} & \textbf{Meaning} & \textbf{Value}  \\ \midrule
           $\eta$ & Parameters that control effect of smooth policy update & 0.25 \vspace{0.25cm} \\
           $\sigma$ & Parameters that control effect of causal entropy & 0.04 \vspace{0.25cm} \\
           $M$ & Number of episodes for one iteration & 10 \vspace{0.25cm} \\
           $T$ & Number of steps for one episode & 300 \vspace{0.25cm} \\
           $J$ & Iterations of discriminator updates & 1 \vspace{0.25cm} \\
           $K$ & Iterations of value network updates & 1 \vspace{0.25cm} \\ \bottomrule
          \end{tabular}
        }
    \end{center}
    \caption{Settings and learning parameters of $2$ DoF manipulator twice-reaching task\label{table:2doftwice_learning_setting}}
\end{table}

\subsubsection{Pick-and-place task}\label{sec:sim_pick}
Figure \ref{fig:pick_and_place} shows an overview of the pick-and-place task. Since this study focuses on learning cloth-manipulation policies for image input, we modified the goal-reaching task \cite{DBLP:journals/corr/abs-1802-09464} into an environment suitable for image input. Since 3D information is unavailable from a single image, the robot hand moves horizontally and only vertically when picking or placing. The environment and the GA-GAIL parameter settings are shown in Tables \ref{table:pick_mdp_setting} and \ref{table:pick_gagail_setting}. 

\begin{table}
    \begin{center}
        \subfloat[Parameter setting of pick-and-place task\label{table:pick_mdp_setting}]{
          \begin{tabular}{@{}p{3cm}p{8cm}l@{}}
          \toprule
          \textbf{MDP setting} & \textbf{Description}  \\ \midrule
          State & Input state is a $84 \times 84$ px color image. Number of state dimensions is $84 \times 84 \times 3 \,(21,168)$. \vspace{0.25cm} \\
          \vspace{-3.1cm}Action & 
          \begin{tabular}{@{}p{8cm}@{}}
          The action space is defined by discrete horizontal movement vectors of the robot hand and the pick-or-place action to grasp or release a block. The direction of horizontal movement is $8$, and the amount of movement is two levels of $[0.7,0.2]$. The pick-or-place action executes a pick when a block is under the hand and a place when the hand is grasping a block. The number of all the actions is $18\,(8\times2+1+1)$, including stop actions.\\
          \begin{minipage}{8cm}
            \centering
            \scalebox{0.6}{\includegraphics[width=1.\columnwidth]{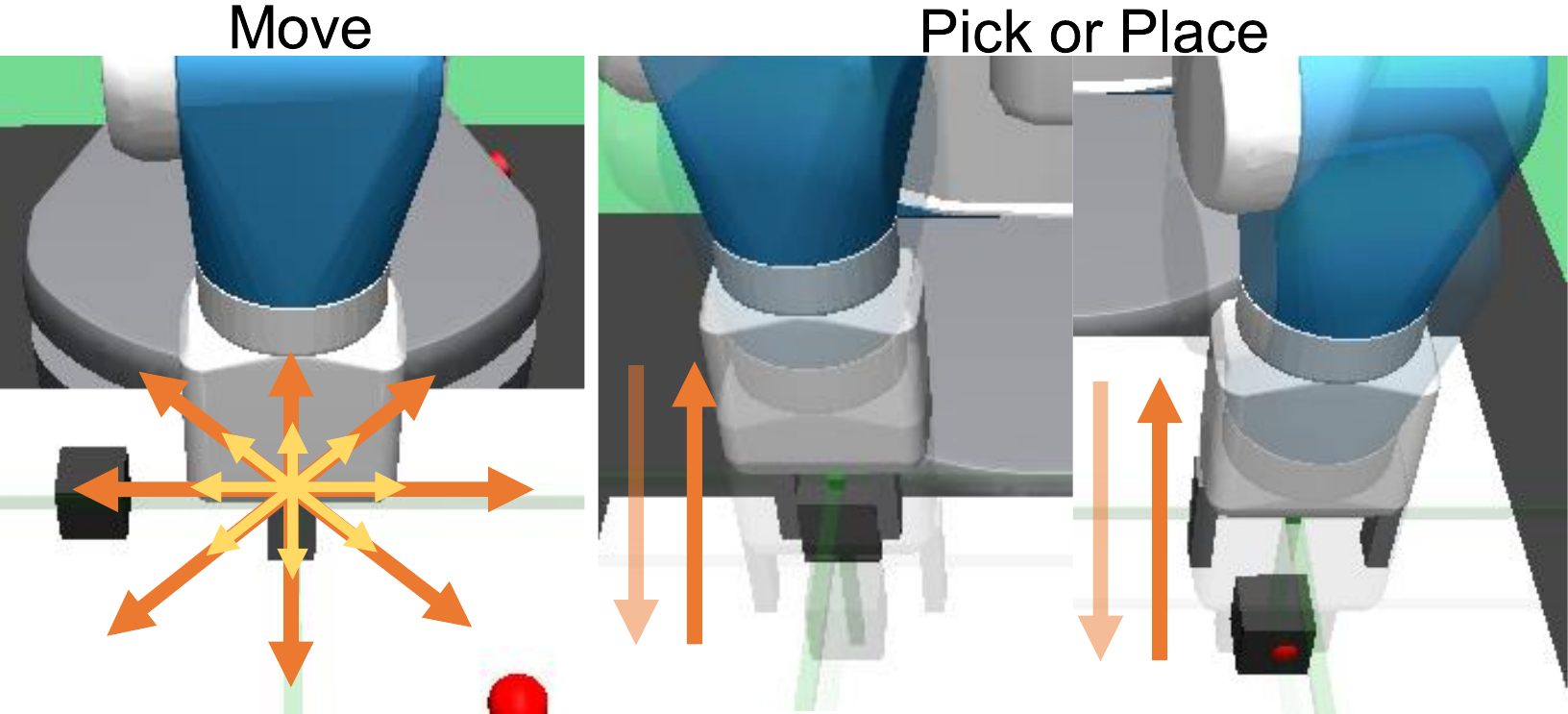}}
          \end{minipage}
          \end{tabular}
          \vspace{0.25cm} \\
          Reward for evaluation & The reward function is the sum of three sub-rewards: reaching reward $r_r$, distance reward $r_d$, and place reward $r_p$. $r_r$ is the distance between the arm and the block, and $r_d$ is the distance between the block and the goal position.  $r_p$ is $[+1,+2]$ when the distance between the block and the goal position is less than $[0.1, 0.05]$ and $0$ otherwise.
          \begin{minipage}{8cm}
            \centering
            \scalebox{0.35}{\includegraphics[width=1.\columnwidth]{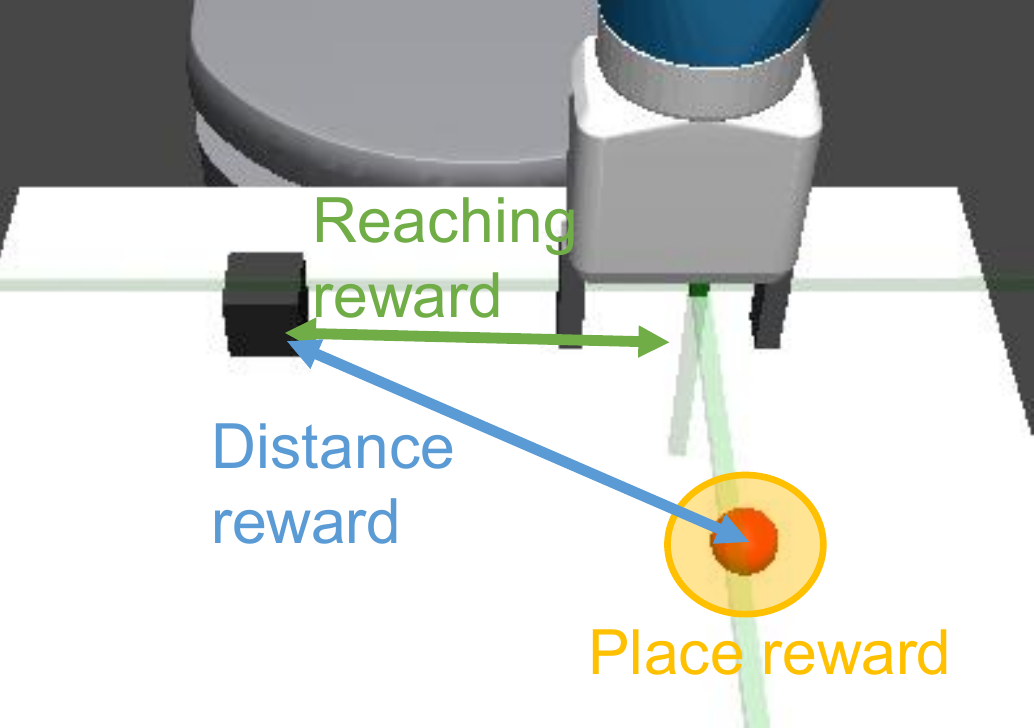}}
          \end{minipage}
          \vspace{0.25cm} \\
          Initial state & Initial state of robot pose, block and goal positions are fixed. \vspace{0.25cm}\\ 
          Demonstrations & Demonstrations are sampled from an RL policy that maximizes the evaluation reward. Demonstrations ($30$ steps $\times$ $5$ trajectory = $150$ samples) are generated from exploring policies during learning process. \vspace{0.25cm}\\
          Goal state & Goal samples are selected from demonstrations, where distance between block and goal position is less than $0.1$.\\
          \bottomrule
          \end{tabular}
        }

    \end{center}
    \caption{Settings and learning parameters of pick-and-place task}
\end{table}

\begin{table}\ContinuedFloat
    \begin{center}
        \subfloat[Parameter setting of GA-GAIL algorithm\label{table:pick_gagail_setting}]{
          \begin{tabular}{@{}lp{8cm}llll@{}}
          \toprule
           \textbf{Parameter} & \textbf{Meaning} & \textbf{Value}  \\ \midrule
           $\eta$ & Parameters that control effect of smooth policy updates & 0.2 \vspace{0.25cm} \\
           $\sigma$ & Parameters that control effect of causal entropy & 0.05 \vspace{0.25cm} \\
           $M$ & Number of episodes for one iteration & 30 \vspace{0.25cm} \\
           $T$ & Number of steps for one episode & 150 \vspace{0.25cm} \\
           $J$ & Iterations of discriminator updates & 1 \vspace{0.25cm} \\
           $K$ & Iterations of value network updates & 1 \vspace{0.25cm} \\ \bottomrule
          \end{tabular}
        }
    \end{center}
    \caption{Settings and learning parameters of pick-and-place task (cont.)\label{table:pick_setting}}
\end{table}

\subsection{Evaluation}
\subsubsection{Learning from imperfect demonstrations}\label{subsec:imperfect}
We first investigated the learning policy performances from imperfect demonstrations. To get different performance demonstrations, we collected demonstrations from RL policies during and after the learning. An imperfect demonstration was defined as a trajectory collected from a during-learning policy that can sometimes achieve the goal state, and a perfect demonstration was defined as a trajectory collected from an after-learning policy that can stably achieve the goal state. We compared EDPN and naive DQN to verify the effect of smooth policy updates and entropy-maximizing. After this step, GAIL with DQN is called GAIL (DQN), and Goal-Aware Generative Adversarial Imitation Learning with DQN is called GA-GAIL (DQN). 

\begin{figure}[th]
    \vspace{1mm}
    \centering
    \begin{tabular}{c}
        \begin{minipage}{0.9\linewidth}
            \centering
            \subfloat[Twice-reaching task]{
            \includegraphics[keepaspectratio, width=0.99\linewidth, angle=0]
                        {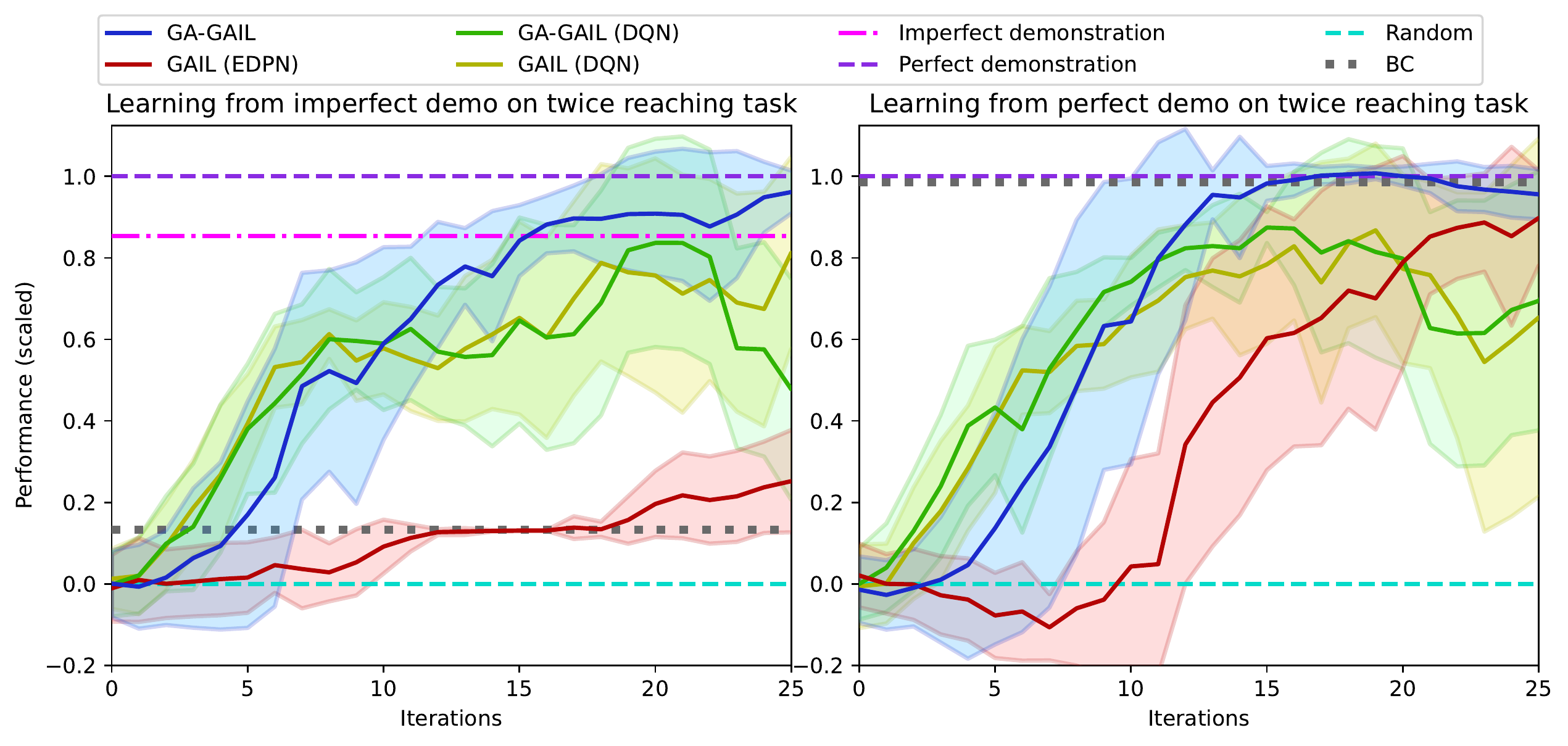}}
        \end{minipage}
        \\
        \begin{minipage}{0.9\linewidth}
            \centering
            \subfloat[Pick-and-place task]{
            \includegraphics[keepaspectratio, width=0.99\linewidth, angle=0]
                        {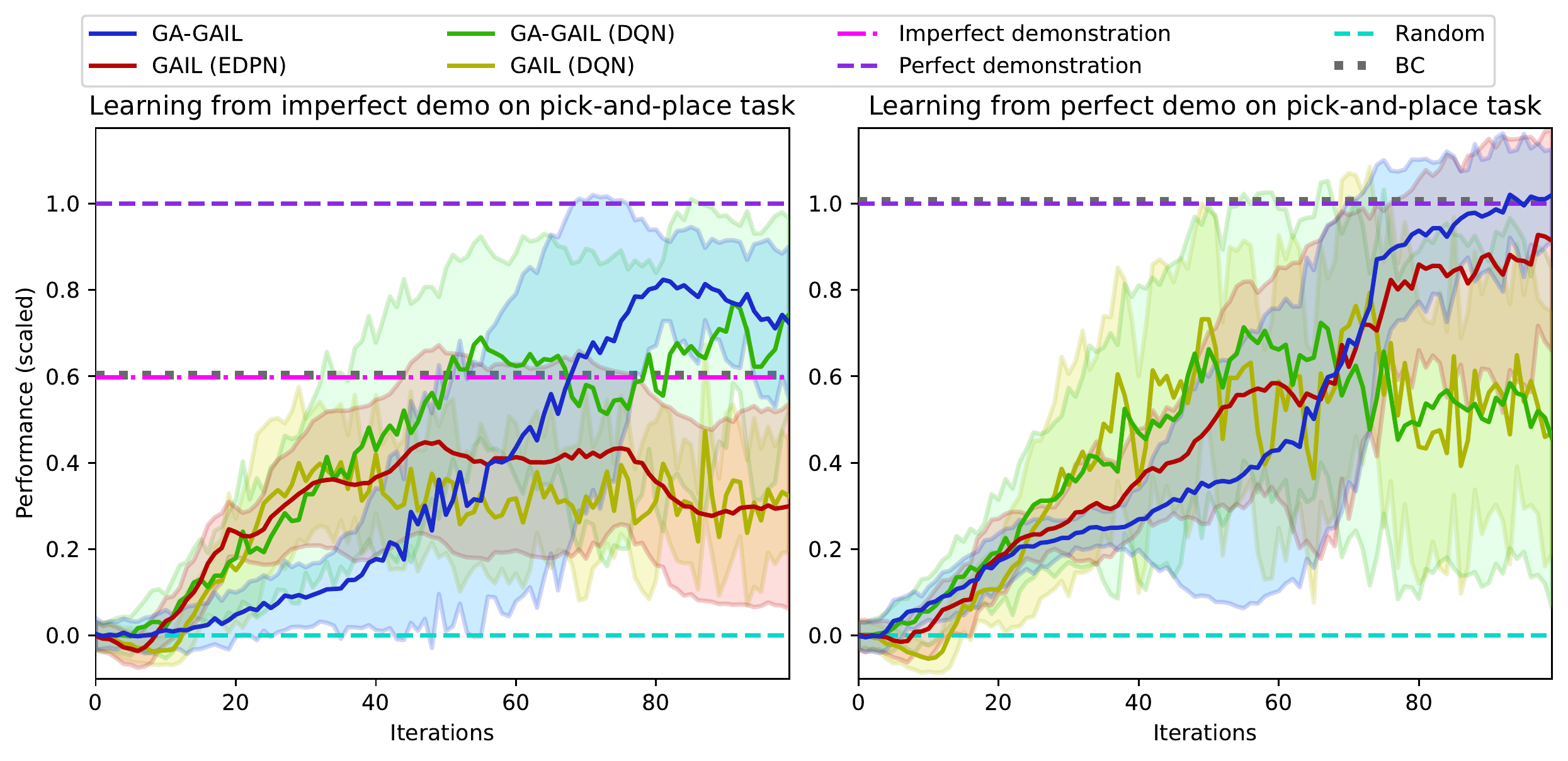}}
        \end{minipage}
    \end{tabular}
    \caption{
        Learning curve for each GAIL framework: Performance is compared by applying each method to (a) twice-reaching task and (b) pick-and-place task. Left graph for each task is learning result from imperfect demonstrations, and the right graph for each task is learning result from perfect demonstrations. Each method's performance on y-axis was scaled by total reward of random policies and demonstrations used for learning. One learning curve shows averaged learning results based on five experiments. Shaded regions of each learning curve represent standard deviation.
    }
    \label{fig:comper_demo}
\end{figure}

The learning results with different demonstrations are shown in Fig. \ref{fig:comper_demo}. The performance of the demonstrations used for learning is scaled to $1$, and one random policy is scaled to 0. The learning curves are the averages of five experiments. In the twice-reaching task, BC's low performance occurred because many imperfect demonstration trajectories failed to achieve the goal state and its reward for the goal state in this task is high. GAIL (EDPN) has high performance when learned from perfect demonstrations but poor performance when learned from imperfect demonstrations. GA-GAIL (DQN) has lower performance than GA-GAIL because it has no constraints to stabilize the learning of policies. Our proposed method GA-GAIL has high performance independent of the task and demonstration. In particular, GA-GAIL policies learned from imperfect demonstrations outperformed the imperfect demonstrations used for learning.

\subsubsection{Comparison with RL from imperfect demonstration}\label{subsec:comper_RL}
Previous subsection compared methods that do not include a task-dependent reward function. In this subsection, we compared the performance of our proposed method with and without the task-dependent reward function. This experiment compared the performance of our proposed method with reinforcement learning (RL) from imperfect demonstrations and task-dependent reward functions. The comparison methods are Normalized Actor-Critic (NAC) \cite{gao2018reinforcement} and Deep Q-learning from Demonstrations (DQfD) \cite{hester2018deep}, both of which initialize the policies from imperfect demonstrations with rewards and update the policies using a task-dependent reward function. DOfD is a standard deep reinforcement learning method with demonstrations. NAC is a deep reinforcement learning method suitable for learning from imperfect demonstrations and reduces the effect of imperfect data in demonstrations by maximizing the entropy of the learning policies. 

\begin{figure}[t]
    \vspace{1mm}
    \centering
    \begin{tabular}{c}
        \hspace{-0.8cm}
        \begin{minipage}{0.5\linewidth}
            \centering
            \subfloat[Twice-reaching task]{
            \includegraphics[keepaspectratio, width=0.99\linewidth, angle=0]
                        {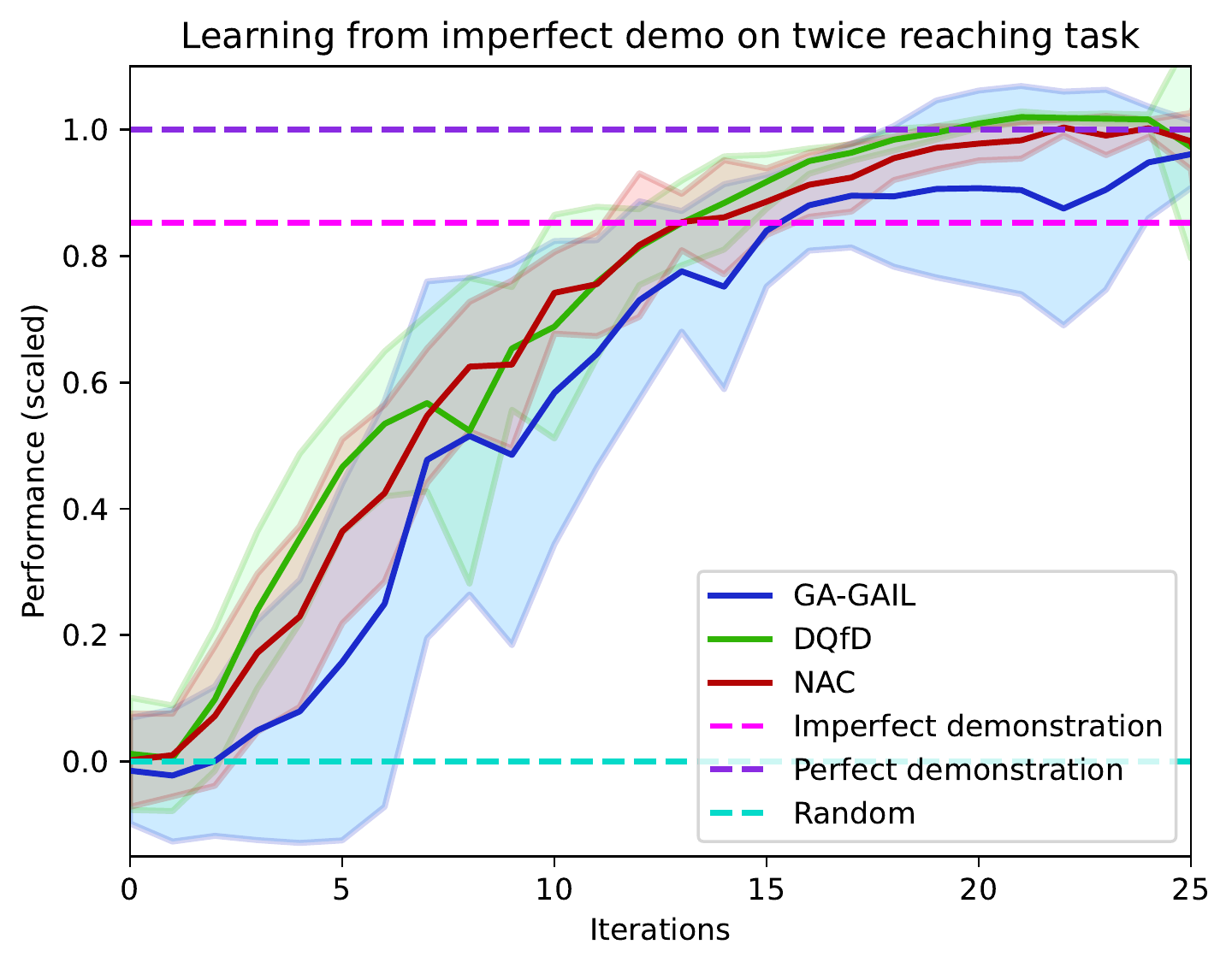}}
        \end{minipage}
        \begin{minipage}{0.5\linewidth}
            \centering
            \subfloat[Pick-and-place task\label{fig:comper_RL_b}]{
            \includegraphics[keepaspectratio, width=0.99\linewidth, angle=0]
                        {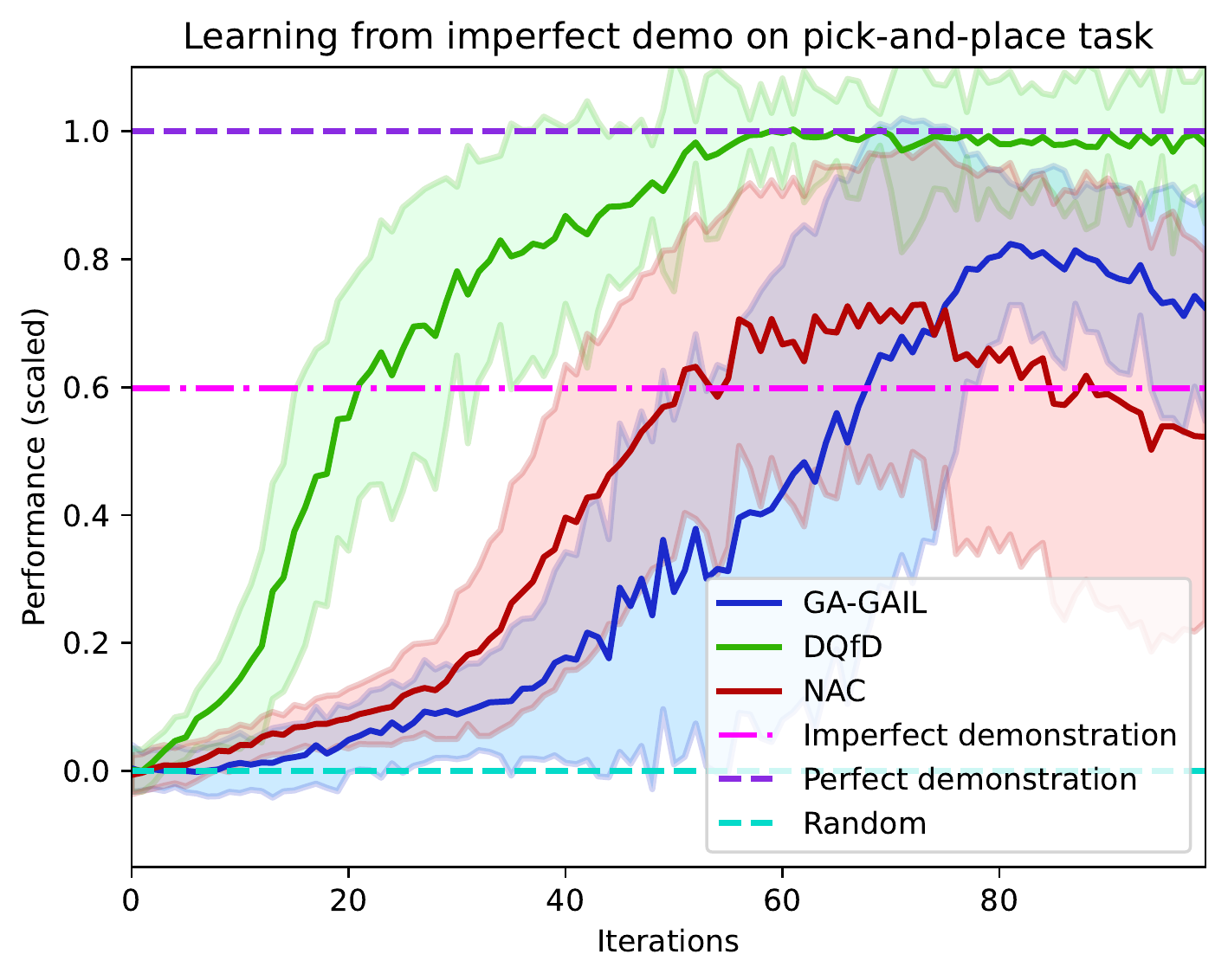}}
        \end{minipage}
    \end{tabular}
    \caption{
        Learning curve of proposed method and RL from imperfect demonstrations: Our proposed method learns from imperfect demonstrations only (with goal labels), and comparison method learns from imperfect demonstrations and task-dependent reward functions. Each method's performance on y-axis was scaled by total reward of random policies and demonstrations used for learning. One learning curve shows averaged learning results based on five experiments. Shaded regions of each learning curve represent standard deviation.
    }
    \label{fig:comper_RL}
\end{figure}

The compared learning results are shown in Fig. \ref{fig:comper_RL}. In the twice-reaching task, even though the proposed method does not use a task-dependent reward function, its performance closely approaches RL methods with imperfect demonstrations and a task-dependent reward function. In the pick-and-place task, the proposed method outperformed NAC whose poor performance can be explained by entropy maximization, which negatively affects the learning due to the high-dimensional action space. In Fig. \ref{fig:comper_RL_b}, GA-GAIL and NAC performance degrade in the middle stage for the following two possible reasons. 1) the pick-and-place task easily degrades performance by exploration actions since it significantly differs in the value functions for optimal and non-optimal actions in executing picking and placing. 2) GA-GAIL and NAC contain constraints that maximize the entropy of the policies encouraging exploration. 

\begin{figure}[t]
    \vspace{1mm}
    \centering
    \begin{tabular}{c}
        \hspace{-0.8cm}
        \begin{minipage}{0.5\linewidth}
            \centering
            \subfloat[Twice-reaching task]{
            \includegraphics[keepaspectratio, width=0.99\linewidth, angle=0]
                        {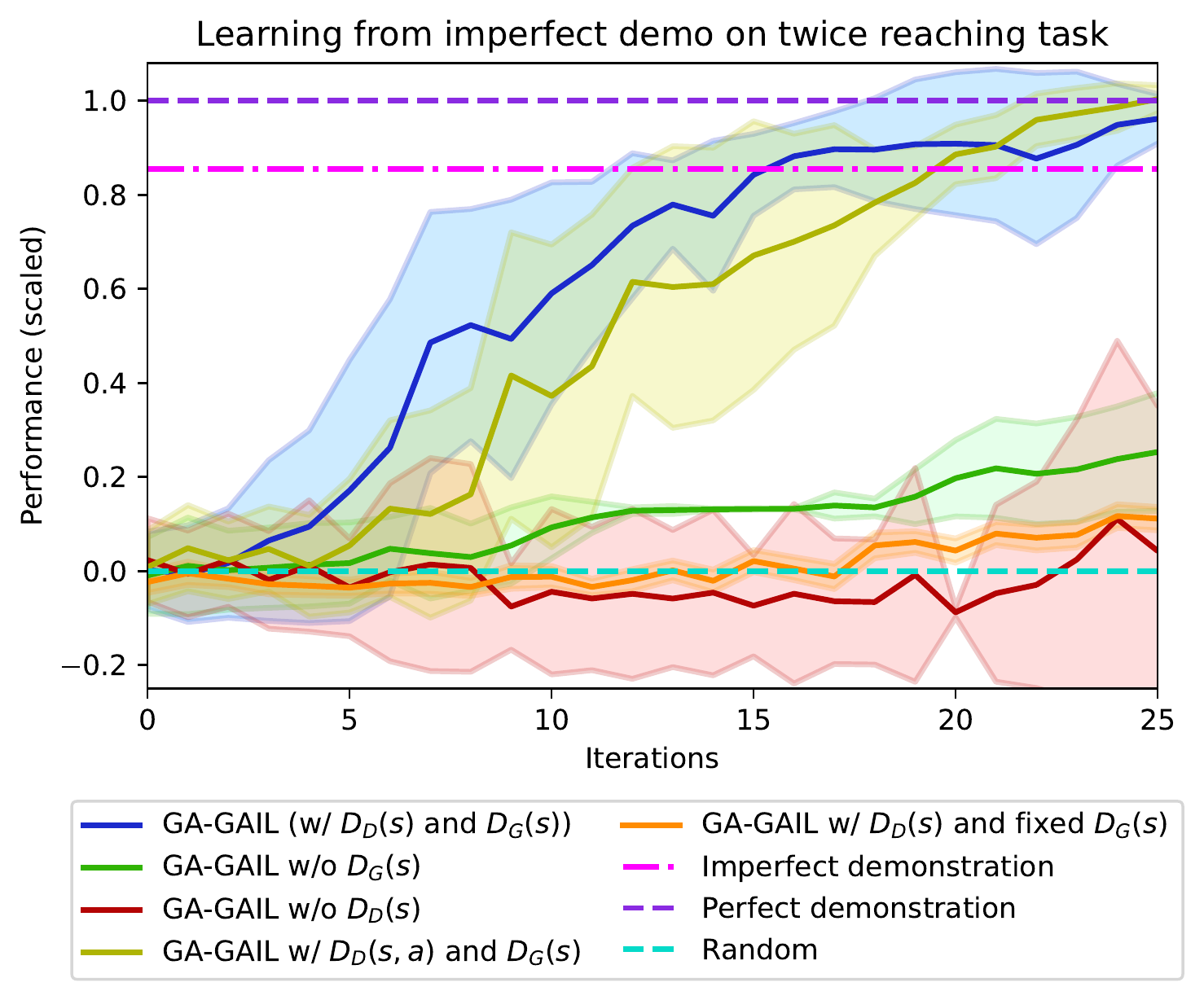}}
        \end{minipage}
        \begin{minipage}{0.5\linewidth}
            \centering
            \subfloat[Pick-and-place task]{
            \includegraphics[keepaspectratio, width=0.99\linewidth, angle=0]
                        {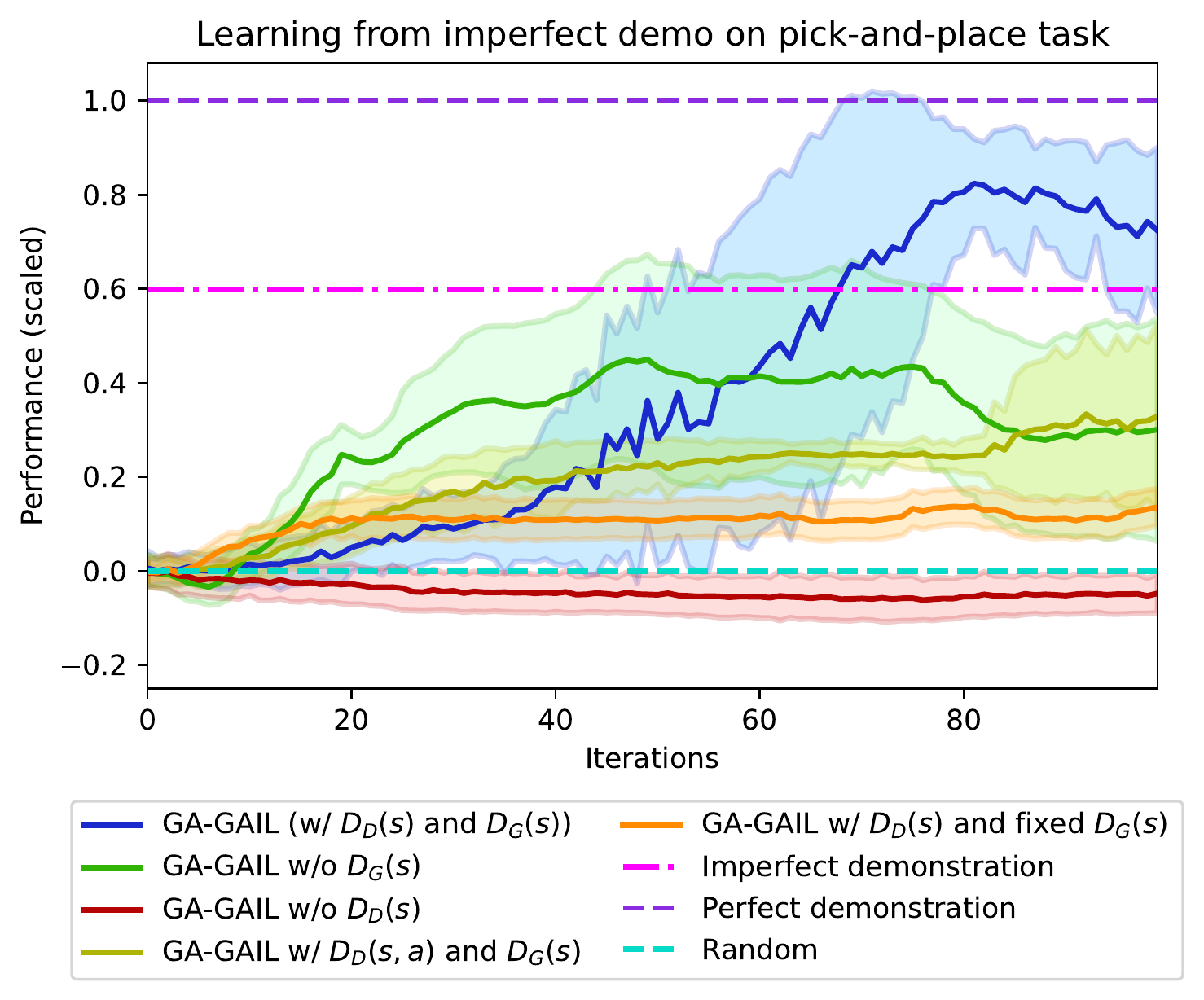}}
        \end{minipage}
    \end{tabular}
    \caption{
        Learning curve for each discriminator setting: Each method's performance on y-axis was scaled by total reward of random policies and demonstrations used for learning. One learning curve shows averaged learning results based on five experiments. Shaded regions of each learning curve represent standard deviation.
    }
    \label{fig:comper_disc}
\end{figure}

\subsubsection{Comparison with different discriminator settings}\label{subsec:disc_set}
We compared the learning results for different discriminator settings on two simulation benchmark tasks to demonstrate that the GA-GAIL discriminator combination is suitable for learning policies. The four discriminator settings to be compared are GA-GAIL w/o $D_G(s)$, GA-GAIL w/o $D_D(s)$, GA-GAIL w/ $D_D(s,a)$ and $D_G(s)$, GA-GAIL w/ $D_D(s)$ and fixed $D_G(s)$. GA-GAIL w/ $D_D(s,a)$ and $D_G(s)$ contains action inputs to the demonstration discriminator $D_D(s,a)$. GA-GAIL w/ $D_D(s)$ and fixed $D_G(s)$ does not learn the goal discriminator adversarially; it learns the goal discriminator as a static binary neural network classifier. The convolutional layer of the two discriminators in Fig. \ref{fig:disc_struct} is shared, but the convolutional layer of $D_D(s)$ and fixed $D_G(s)$ is separate. Updating $D_D(s)$ does not change the output of the fixed $D_G(s)$, which was pre-trained from goal labels and $4500$ data collected from a random policy.

\begin{figure}[t]
    \centering
    \begin{tabular}{c}
        \hspace{-0.8cm}
        \begin{minipage}{0.5\linewidth}
            \centering
            \subfloat[Twice-reaching task]{
            \includegraphics[keepaspectratio, width=0.95\linewidth, angle=0]
                        {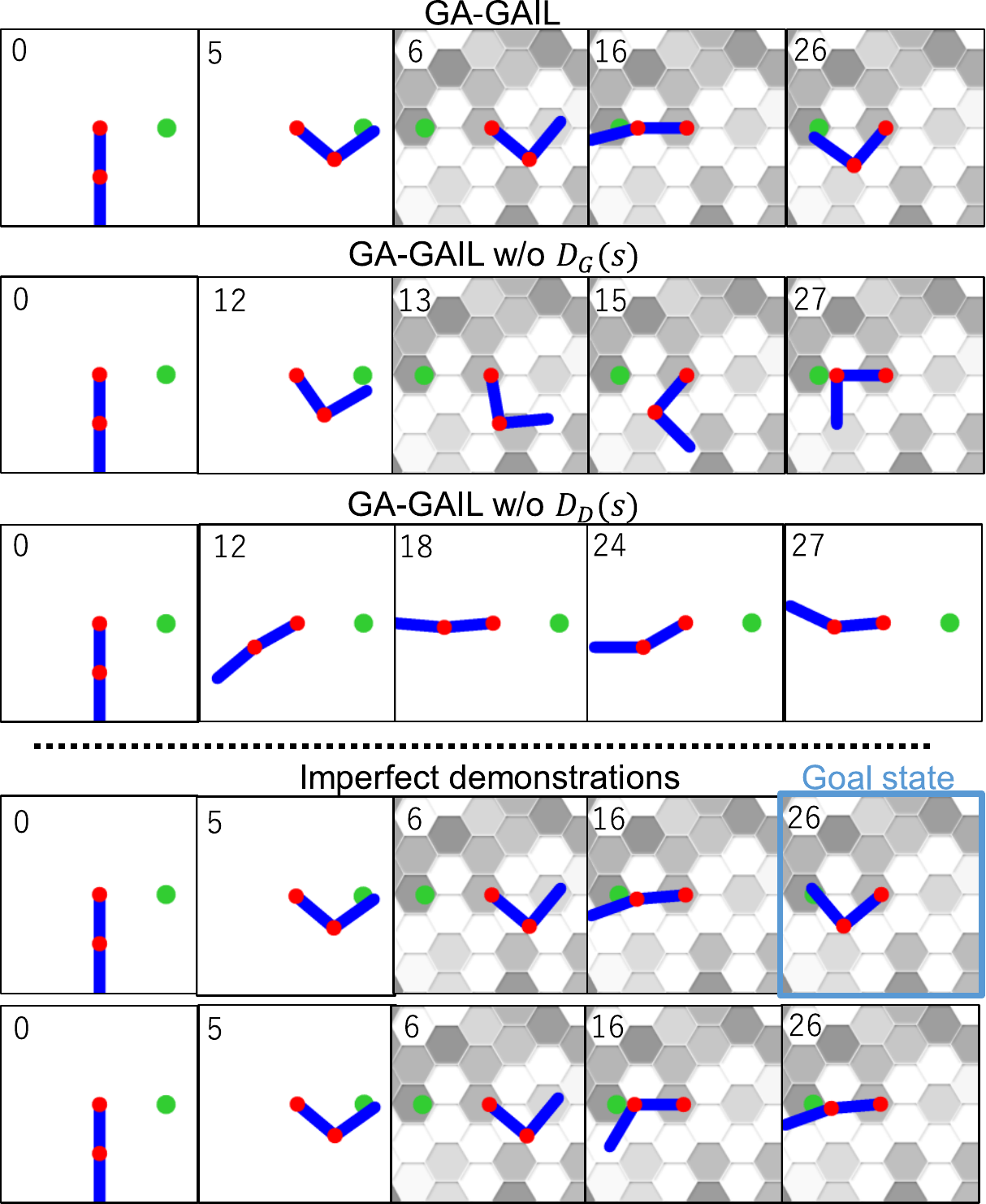}}
        \end{minipage}
        \begin{minipage}{0.5\linewidth}
            \centering
            \subfloat[Pick-and-place task]{
            \includegraphics[keepaspectratio, width=0.99\linewidth, angle=0]
                        {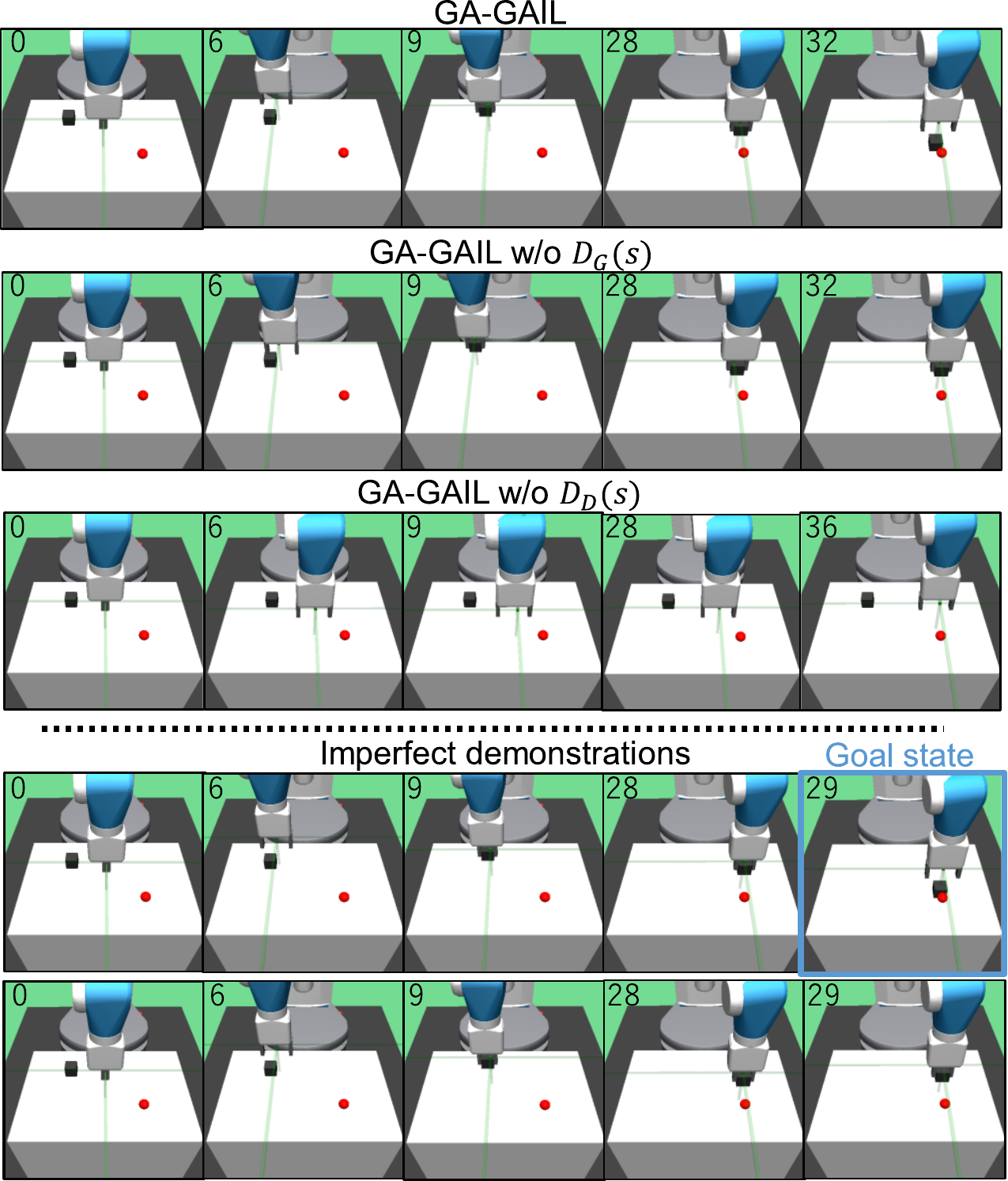}}
        \end{minipage}
    \end{tabular}
    \caption{Snapshot of trajectories generated from learned policies: Number in upper left corner indicates steps. Blue frame in demonstrations shows labeled goal state.}
    \label{fig:snapshot_trajectory}
\end{figure}

Figure \ref{fig:comper_disc} shows the learning results and compares each discriminator setting, and Fig. \ref{fig:snapshot_trajectory} shows the learned trajectories for some methods. The GA-GAIL w/o $D_G(s)$ performance is low because it is learning from imperfect demonstrations without goal discriminator $D_G(s)$. GA-GAIL w/o $D_D(s)$ failed to learn because no demonstration discriminator $D_D(s)$ guides the trajectory to the goal state. Since the reward based on the goal discriminator is concerned with discriminating between goal states and others, its contribution to states far from the goal may be limited. Thus, GA-GAIL uses it in conjunction with rewards based on an imperfect-demonstration discriminator, where The goal discriminator works with the focus around the goal state complementary to the imperfect-demonstration discriminator, which works widely in state space. Such a combination performs better than only one of them is used alone as shown in Fig. \ref{fig:comper_disc}. 

GA-GAIL w/ $D_D(s,a)$ and $D_G(s)$ performed well on the twice-reaching task but poorly on the pick-and-place task. GA-GAIL w/ $D_D(s)$ and fixed $D_G(s)$ has low performance for two possible reasons: 1) a limit to the type of data collected with a random policy, and 2) the objective of the learning policy for generating a state that tricks the two discriminators. Since the fixed $D_G(s)$ is not updated from the state generated from the learning policy, GA-GAIL w/ $D_D(s)$ and fixed $D_G(s)$ estimated the reward as high with the incorrect state as the goal state. GA-GAIL outperformed the imperfect demonstration in both tasks, and a combination of discriminators in the proposed method is best suited for both tasks.

\subsubsection{Comparison with different goal label selection methods}\label{subsec:goal_label}
GA-GAIL uses goal labels as additional information compared to GAIL and RL. In this subsection, we investigated the impact of the goal label selection method on the proposed method. We compared the following three goal label selection methods: 1) selecting all the goal states, 2) selecting some of the last two frames of the best performing demonstration trajectories, and 3) selecting one of the best goal states. Since rewards are available for comparison by the simulation task, each goal label selection method is based on rewards. 

\begin{figure}[t]
    \vspace{1mm}
    \centering
    \begin{tabular}{c}
        \hspace{-0.8cm}
        \begin{minipage}{0.5\linewidth}
            \centering
            \subfloat[Twice-reaching task]{
            \includegraphics[keepaspectratio, width=0.99\linewidth, angle=0]
                        {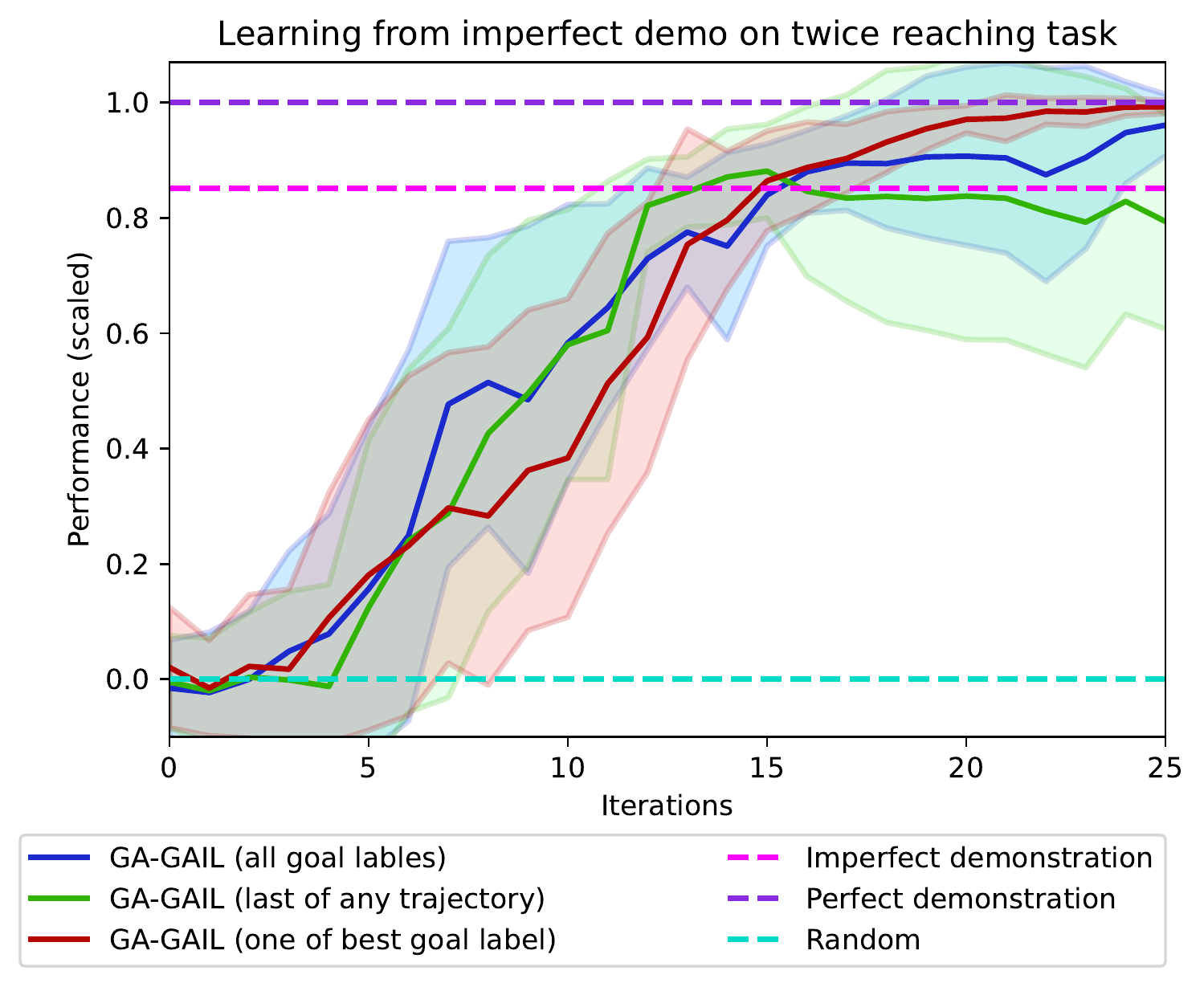}}
        \end{minipage}
        \begin{minipage}{0.5\linewidth}
            \centering
            \subfloat[Pick-and-place task]{
            \includegraphics[keepaspectratio, width=0.99\linewidth, angle=0]
                        {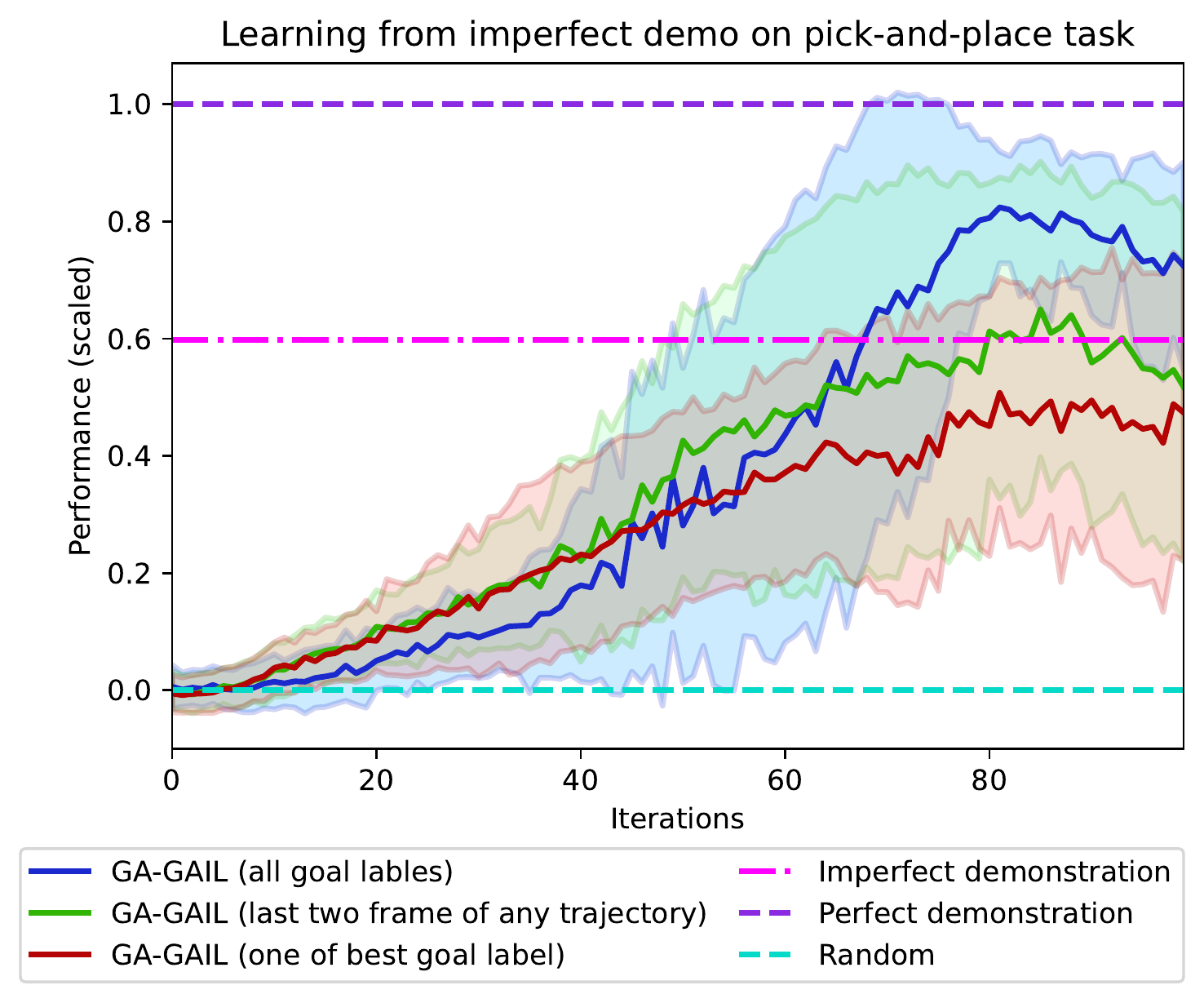}}
        \end{minipage}
    \end{tabular}
    \caption{
        Learning curve for each goal label selection method: Each method's performance on y-axis was scaled by the total reward of random policies and demonstrations used for learning. One learning curve shows averaged learning results based on five experiments. Shaded regions of each learning curve represent standard deviation.
    }
    \label{fig:comper_label}
\end{figure}

The learning results compared are shown in Fig. \ref{fig:comper_label}. GA-GAIL (one of best goal label) has the best performance in the twice-reaching task, but the worst performance in the pick-and-place task. GA-GAIL (last of any trajectory) has a high variance in performance at the end of the learning in both tasks, and its learning is unstable. The performance of GA-GAIL (all goal labels) is stably higher than the other goal state selections in both tasks. 

\section{Cloth-manipulation task with a real robot}\label{sec:real}
In this section, we applied GA-GAIL to NEXTAGE \footnote{www.nextage.kawada.jp/en/}, a $15$-DoF humanoid robot with sufficient precision for complex manufacturing tasks, to learn a policy that turns a handkerchief over and folds clothing from a human demonstration. We focused on learning a policy with high-level discrete actions, i.e., grasp-release points and folding lines on a cloth to solve two tasks: 1) turning a handkerchief over and 2) folding a shirt and a pair of shorts. Since designing a reward function for folding a shirt and a pair of shorts is especially challenging, it is suitable for demonstrating our proposed method without a task-specific reward function design. Video of the real experiments are available at \url{https://youtu.be/h_nII2ooUrE}.

\subsection{Turning a handkerchief over}
\subsubsection{Setting}
\begin{figure}
  \centering
  \subfloat[Recognize cloth and enclose it in a square\label{fig:range}]{
      \includegraphics[width=0.25\columnwidth]{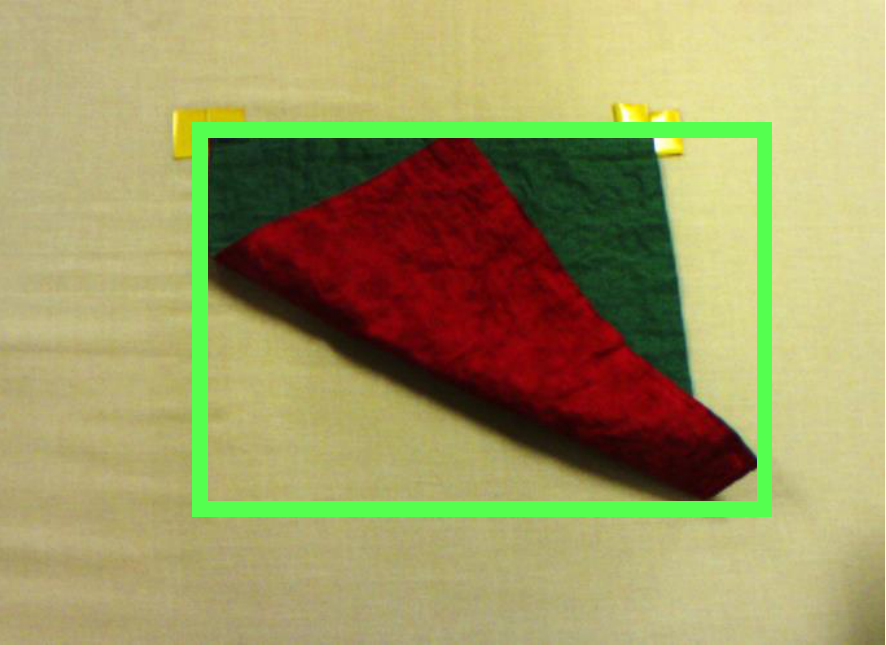}
  }
  \hspace{3mm}
  \subfloat[Divide each side at equal intervals\label{fig:divide}]{
      \includegraphics[width=0.25\columnwidth]{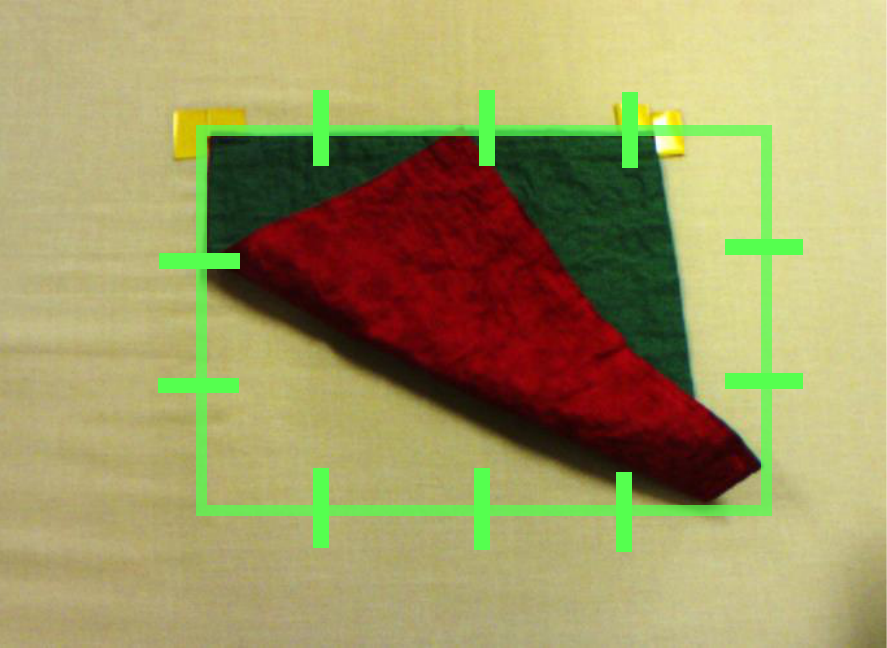}
  }
  \hspace{3mm}
  \subfloat[Scan its edge\label{fig:scanning}]{
      \includegraphics[width=0.25\columnwidth]{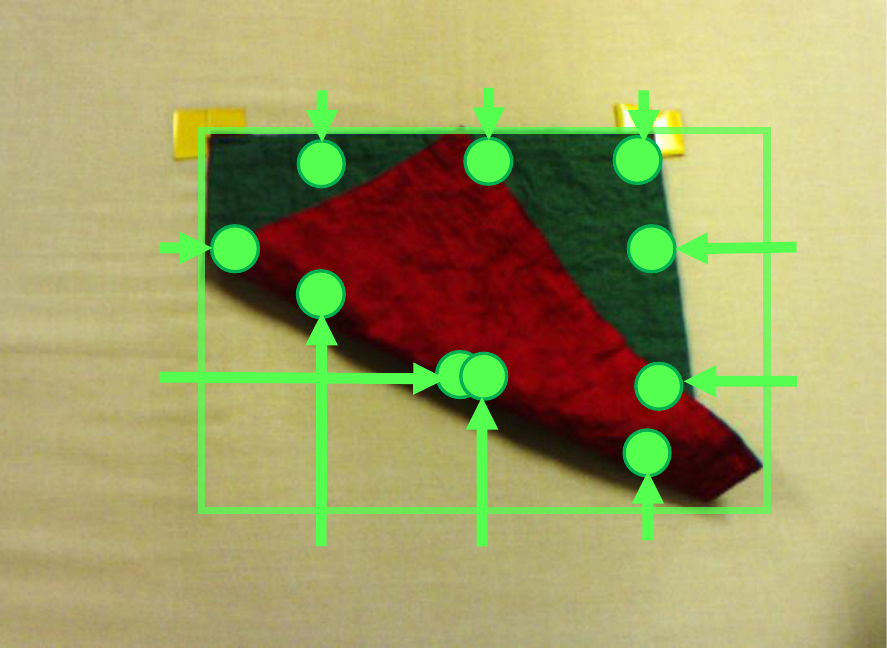}
  }
  \caption{Image processing of picking up points}
  \label{fig:picking_up}
\end{figure}

In this handkerchief task, we evaluated the learned policy's performance in a real-world environment with a designed reward function. This handkerchief has different colors on its back and front, and the goal state is showing the red color spreading on up. The initialization of the handkerchief state is a random pick-and-release operation repeated three times. Since handkerchief initialization is performed by the robot, the environment is suitable for long-term learning. The discriminator settings for GA-GAIL are identical as in Section \ref{subsec:disc_para}. This task's environment and GA-GAIL settings are shown in Tables \ref{table:flip_env_setting} and \ref{table:flip_GAIL_setting}. $10 \times 6 = 60$ gripper actions are defined as picking up the handkerchief from $10$ points over its current area and dropping it to $2 \times 3$ points over the table (Fig. \ref{fig:picking_up}). In this actual experiment, the proposed method was compared with three others: GAIL (EDPN), GA-GAIL (DQN), and GAIL (DQN). GAIL (EDPN) updates policies with EDPN in the standard GAIL framework, GA-GAIL (DQN) updates policies with DQN from reward functions including goal discriminators, and GAIL updates policies with DQN in the standard GAIL framework.

\begin{table}
    \begin{center}
        \subfloat[Parameter setting of turning-handkerchief-over task\label{table:flip_env_setting}]{
          \begin{tabular}{@{}p{3cm}p{8cm}l@{}}
          \toprule
          \textbf{MDP setting} & \textbf{Description}  \\ \midrule
          State & Input state is a $84 \times 84$ px RGB image from NEXTAGE's integrated camera. \vspace{0.25cm} \\
          \vspace{-2cm}Action & 
          \begin{tabular}{@{}p{8cm}@{}}
          $10\times6=60$ gripper actions are defined as picking handkerchief up from 10 points over its current area and dropping it to $2\times3$ points over table (Fig. \ref{fig:picking_up}).\\
          \begin{minipage}{8cm}
            \centering
            \scalebox{0.7}{\includegraphics[width=1.\columnwidth]{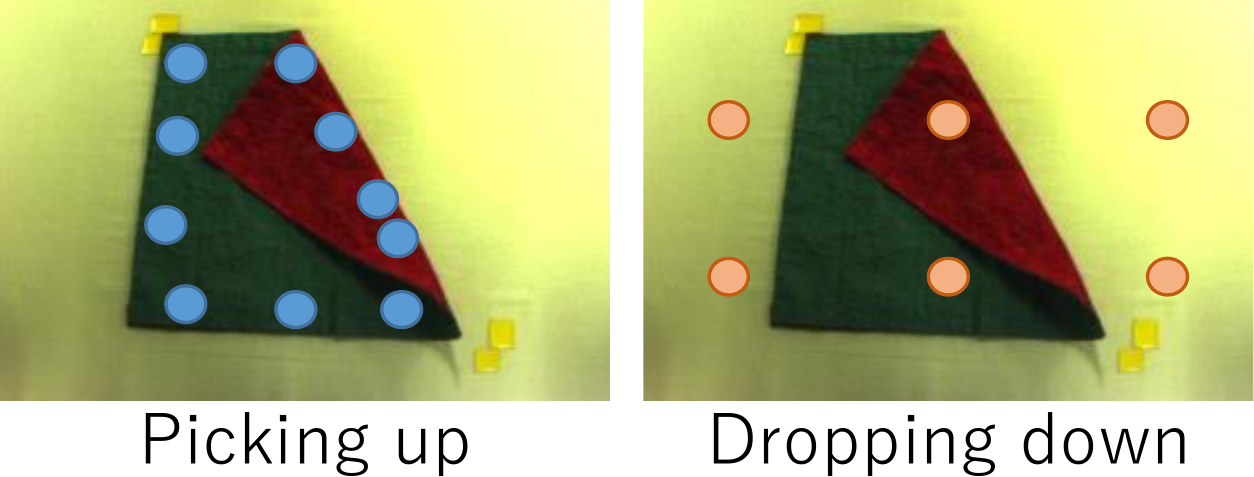}}
          \end{minipage}
          \end{tabular}
          \vspace{0.25cm} \\
          Reward & Reward is defined as ratio of red area over the whole image in current state.
          \begin{minipage}{8cm}
            \centering
            \scalebox{0.35}{\includegraphics[width=1.\columnwidth]{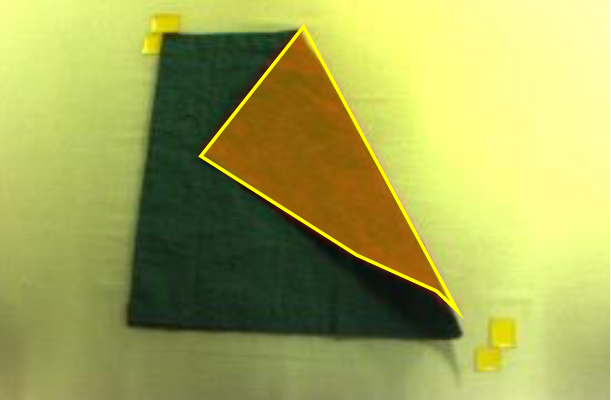}}
          \end{minipage}
          \vspace{0.25cm} \\
          Initial state & Robot executes three random actions and initializes handkerchief. \vspace{0.25cm}\\ 
          Demonstrations & Samples are collected from humans for trajectories of $5$ episodes ($100$ samples). \vspace{0.25cm}\\
          Goal state & Goal state is set to when red area finally exceeds $80\%$.\\
          \bottomrule
          \end{tabular}
        }

        \subfloat[Parameter setting of GA-GAIL algorithm\label{table:flip_GAIL_setting}]{
          \begin{tabular}{@{}lp{8cm}llll@{}}
          \toprule
           \textbf{Parameter} & \textbf{Meaning} & \textbf{Value}  \\ \midrule
           $\eta$ & Parameters that control effect of smooth policy updates & 1 \vspace{0.25cm} \\
           $\sigma$ & Parameters that control effect of causal entropy & 0.05 \vspace{0.25cm} \\
           $M$ & Number of episodes for one iteration & 10 \vspace{0.25cm} \\
           $T$ & Number of steps for one episode & 30 \vspace{0.25cm} \\
           $J$ & Iterations of discriminator updates & 5 \vspace{0.25cm} \\
           $K$ & Iterations of value network updates & 10 \vspace{0.25cm} \\ \bottomrule
          \end{tabular}
        }
    \end{center}
    \caption{Settings and learning parameters of turning-handkerchief-over task\label{table:flip_learning_setting}}
\end{table}

\subsubsection{Results}

\begin{figure}
  \begin{center}
    \includegraphics[clip,scale=0.6]{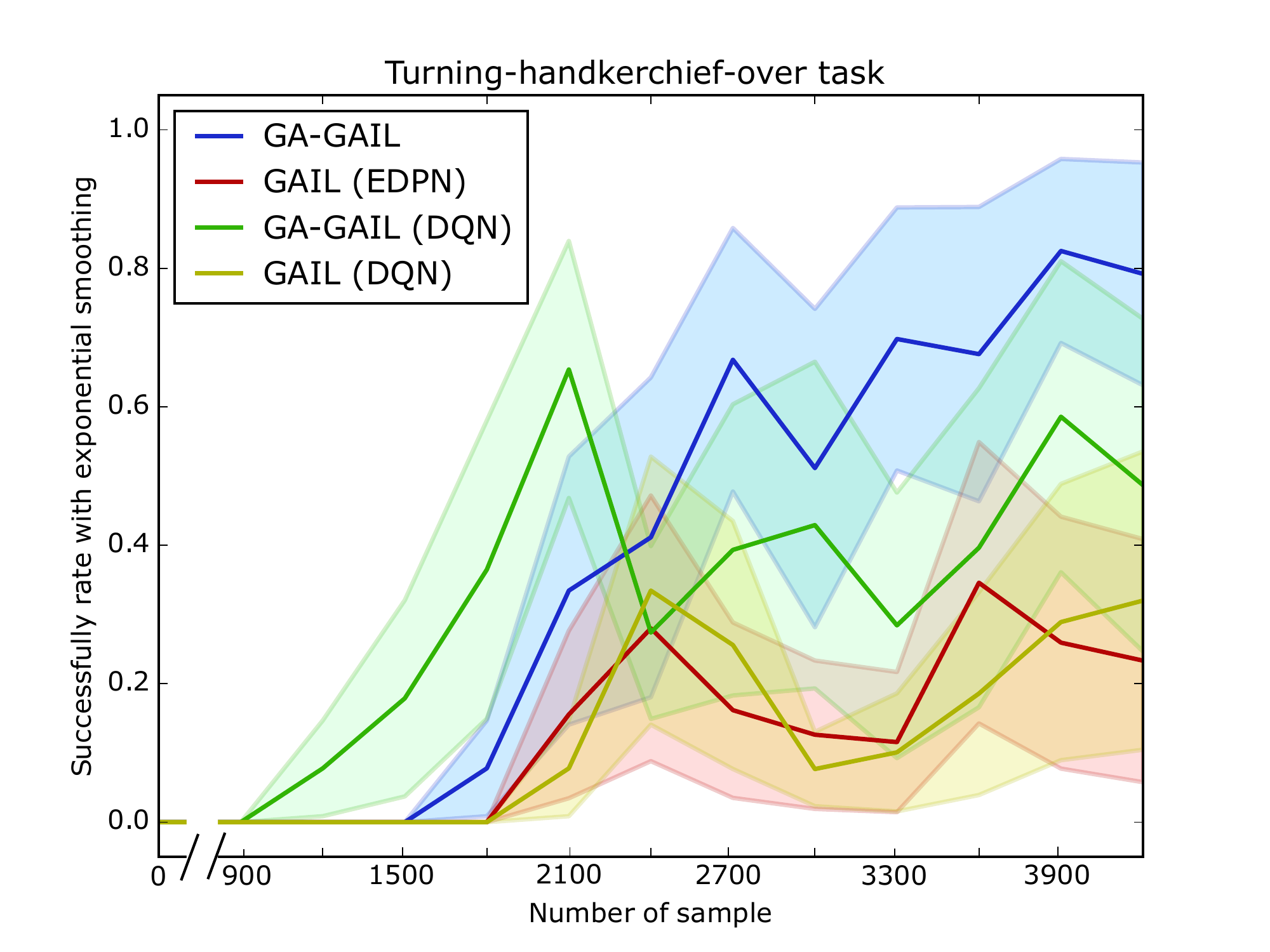}
    \caption{Learning results of NEXTAGE humanoid robot on turning-handkerchief-over task: GA-GAIL, GAIL (EDPN), GA-GAIL (DQN) and GAIL (DQN) were compared on learning curves evaluated by task success rate. Each method learns policy $3$ times, and each policy is tried for $3$ episodes. Thus, each method was evaluated by task success rate from $9\,(3 \times 3)$ episodes. Shaded regions of each learning curve represent standard deviation.}
    \label{fig:flip_results}
  \end{center}
\end{figure}

\begin{figure*}
  \begin{center}
    \includegraphics[width=1.0\columnwidth]{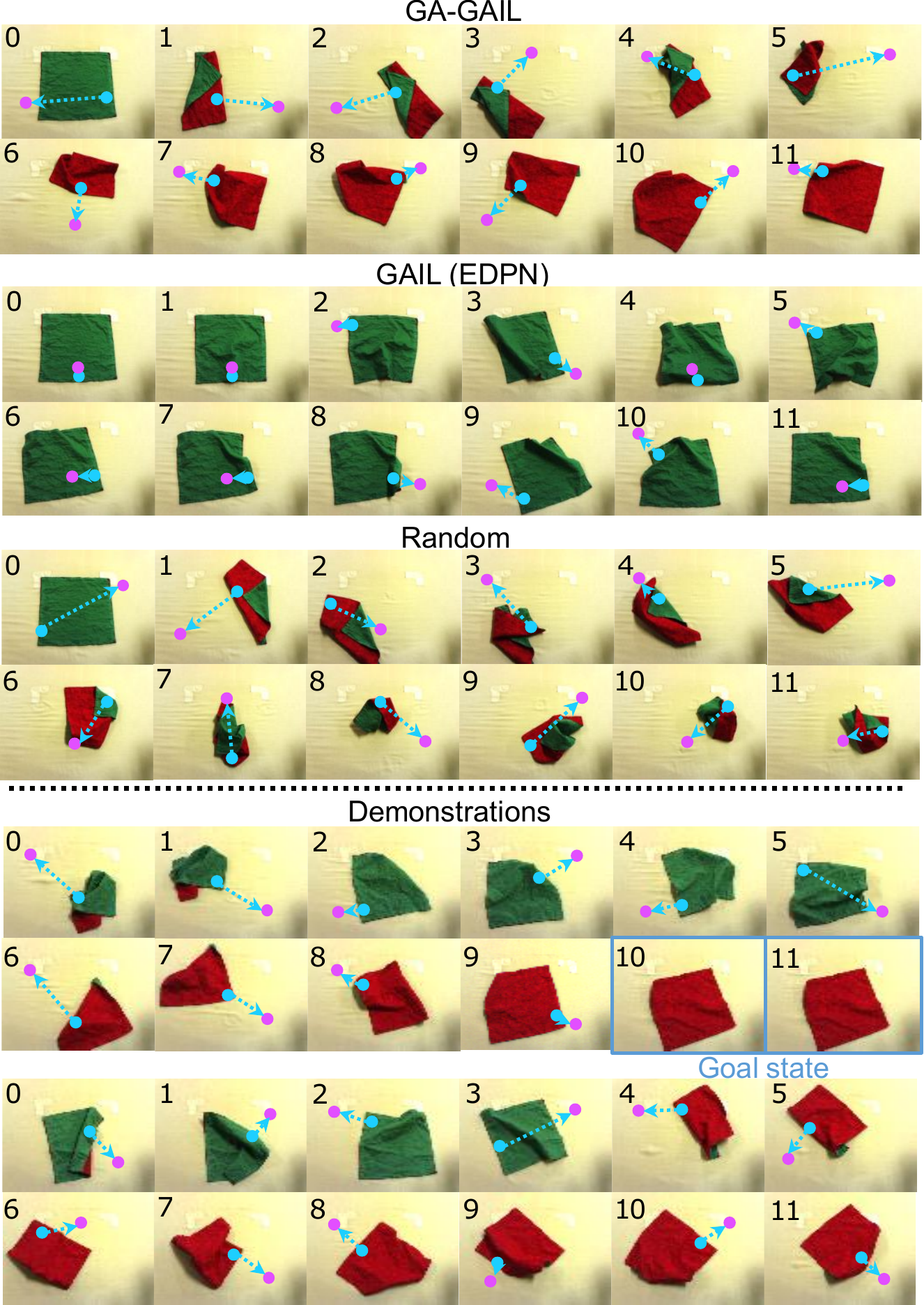}
    \caption{Snapshot of learned policy per step: Blue and purple dots are picking up and dropping positions for actions at each step. Blue frames in demonstrations show labeled goal state.}
    \label{fig:flip_snapshot}
  \end{center}
\end{figure*}

The learning results of each method over three experiments are shown in Fig. \ref{fig:flip_results} where the y-axis is the success rate.
During the test, the handkerchief is initialized to a green spreading up, and we tested three trials per policy ($3$ experiments $\times$ $3$ trials $= 9$ evaluations). We defined a successful manipulation as when the red area finally exceeds $80\%$. Each experiment took about eight hours, including 47 minutes for automatically initializing the handkerchief ($\approx 20$ seconds per episode) to generate $4200$ samples ($=14$ iterations). GAIL (EDPN) and GAIL (DQN) without the goal discriminator, which had roughly the same success rate, were not highly successful. GA-GAIL (DQN) progressed quickly, but its learning is unstable because it does not include constraints to stabilize the policy updates. On the other hand, our proposed method (GA-GAI) has stable learning, reaching a success rate of $80\%$ after $3900$ samples.

Compared to the control policies learned by GA-GAIL and GAIL (EDPN), and the random policies shown in Fig. \ref{fig:flip_snapshot}, the proposed method generated a trajectory that successfully turned the handkerchief over. GAIL (EDPN) kept the green state and failed to turn it over because the demonstrations contained a state that resembles the initial state.

\subsubsection{Evaluation of learned policy's robustness}
We evaluated the robustness of the policies learned by GA-GAIL with different initial states (Table \ref{table:flip_init_test}).
For each initial state, we tested the policy with five trials. We defined a successful manipulation as when the red area finally exceeds $80\%$. According to the result in Table \ref{table:flip_init_test}, Behavioral Cloning (BC) could not select an appropriate action when the observed initial states did not exist in the demonstrations. GA-GAIL (DQN) includes a goal discriminator, although its success rate varies widely, depending on the initial state due to the effect of unstable policy learning. GA-GAIL shows a generalization ability to maximize the red area in various initial states, indicating that it can smoothly explore the turning-over task following the demonstrations.

\begin{table}
\scalebox{0.925}{
  \centering
  \begin{tabular}{cC{1.8cm}C{1.75cm}C{1.75cm}C{1.5cm}C{1.5cm}}
    \toprule
      \textbf{Initial state} & \textbf{GA-GAIL} &\begin{tabular}{c}\textbf{\!\!\!GA-GAIL}\\\textbf{\!\!\!(DQN)}\end{tabular} & \begin{tabular}{c}\textbf{\!GAIL}\\\textbf{\!(EDPN)}\end{tabular} & \textbf{BC} & \begin{tabular}{c}\textbf{\!\!\!\!\!Imperfect}\\\textbf{\!\!\!\!\!demo}\end{tabular} \qquad  \\ \midrule
      \begin{minipage}{2cm}
      \centering
      \scalebox{0.8}{\includegraphics[width=1\columnwidth]{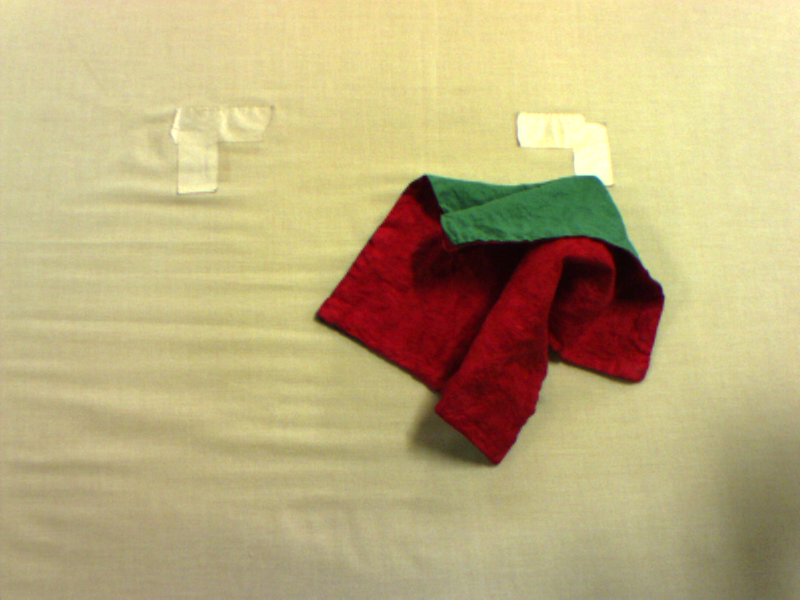}}\\
      \vspace{-0.1cm}
      Random
      \end{minipage}
      & {\Large $\mathbf{4/5}$} & {\Large $3/5$} & {\Large $3/5$} & {\Large $2/5$} & {\Large $3/5$} \vspace{0.7cm} \\
      \begin{minipage}{2cm}
      \centering
      \scalebox{0.8}{\includegraphics[width=1\columnwidth]{./fig/real_robo/flip/init_state/hanker_init0.png}}
      \end{minipage}
      & \textbf{\Large $\mathbf{4/5}$}  & {\Large $2/5$} & {\Large $1/5$} & {\Large $\mathbf{4/5}$} & {\Large ---} \vspace{0.5cm} \\ 
      \begin{minipage}{2cm}
      \centering
      \scalebox{0.8}{\includegraphics[width=1\columnwidth]{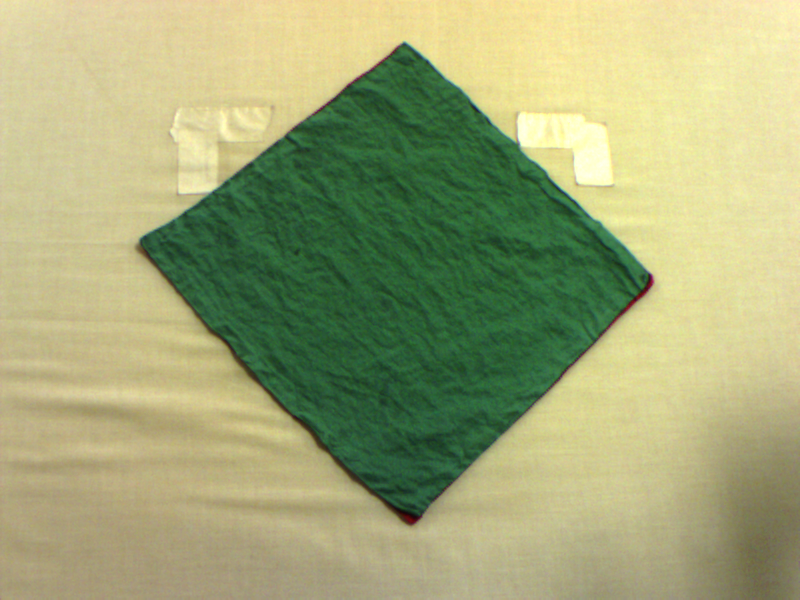}}
      \end{minipage}
      & \textbf{\Large $3/5$} & {\Large $\mathbf{4/5}$} & {\Large $2/5$} & {\Large $2/5$} & {\Large ---} \\ \bottomrule
  \end{tabular}
 \caption{Evaluation of success rate in turning-handkerchief-over task: Each method was evaluated with $5$ trials in each initial state. The imperfect demonstration success rate was calculated using the demonstration trajectories used for learning.}\label{table:flip_init_test}
}
\end{table}

\subsection{Folding a shirt and a pair of shorts}

\subsubsection{Setting}
\begin{figure*}
  \begin{center}
    \includegraphics[width=1.0\columnwidth]{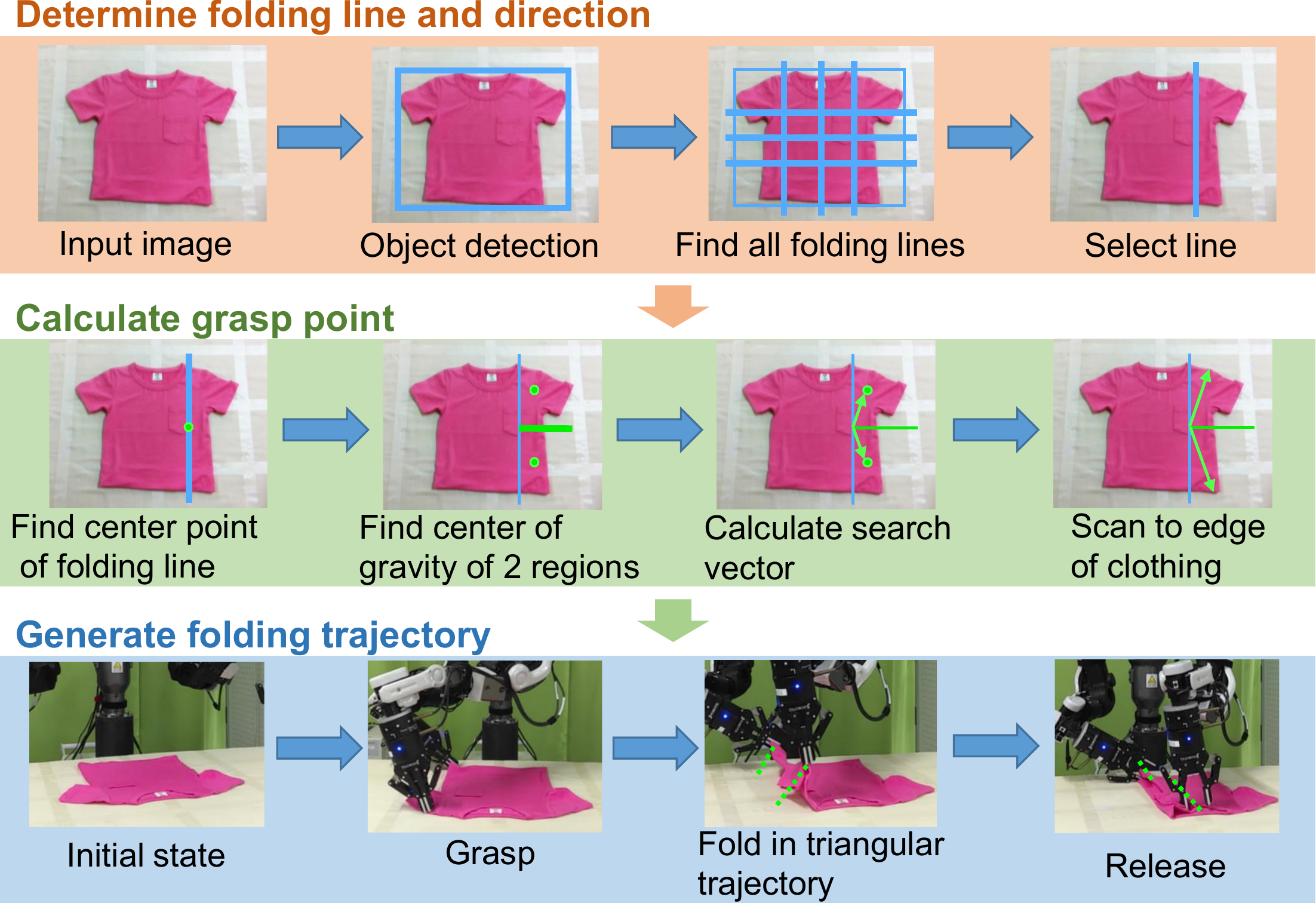}
    \caption{Overview of folding action}
    \label{fig:folding_action}
  \end{center}
\end{figure*}

To verify that the proposed method can learn to manipulate different clothing in the same environment setting, we applied it to folding tasks for shirts and shorts. The discriminator settings for GA-GAIL are identical as in Section \ref{subsec:disc_para}. The environment and the GA-GAIL settings of this task are shown in Tables \ref{table:fold_env_setting} and \ref{table:fold_GAIL_setting}. The folding action is shown in Fig. \ref{fig:folding_action}. The folding direction, grasping points, and folding path are automatically calculated based on the current shape of the article of clothing and the selected fold line.

\begin{table}
    \begin{center}
        \subfloat[Parameter setting for turning-handkerchief-over task\label{table:fold_env_setting}]{
          \begin{tabular}{@{}p{3cm}p{8cm}l@{}}
          \toprule
          \textbf{MDP setting} & \textbf{Description}  \\ \midrule
          State & Input state is a $84 \times 84$ px RGB image from a realsense camera. \vspace{0.25cm} \\
          Action & All fold lines are defined according to shape of current article of clothing ($3+3=6$ line). Fold direction is fixed to move toward center.
          \begin{minipage}{8cm}
            \centering
            \scalebox{0.7}{\includegraphics[width=0.6\columnwidth]{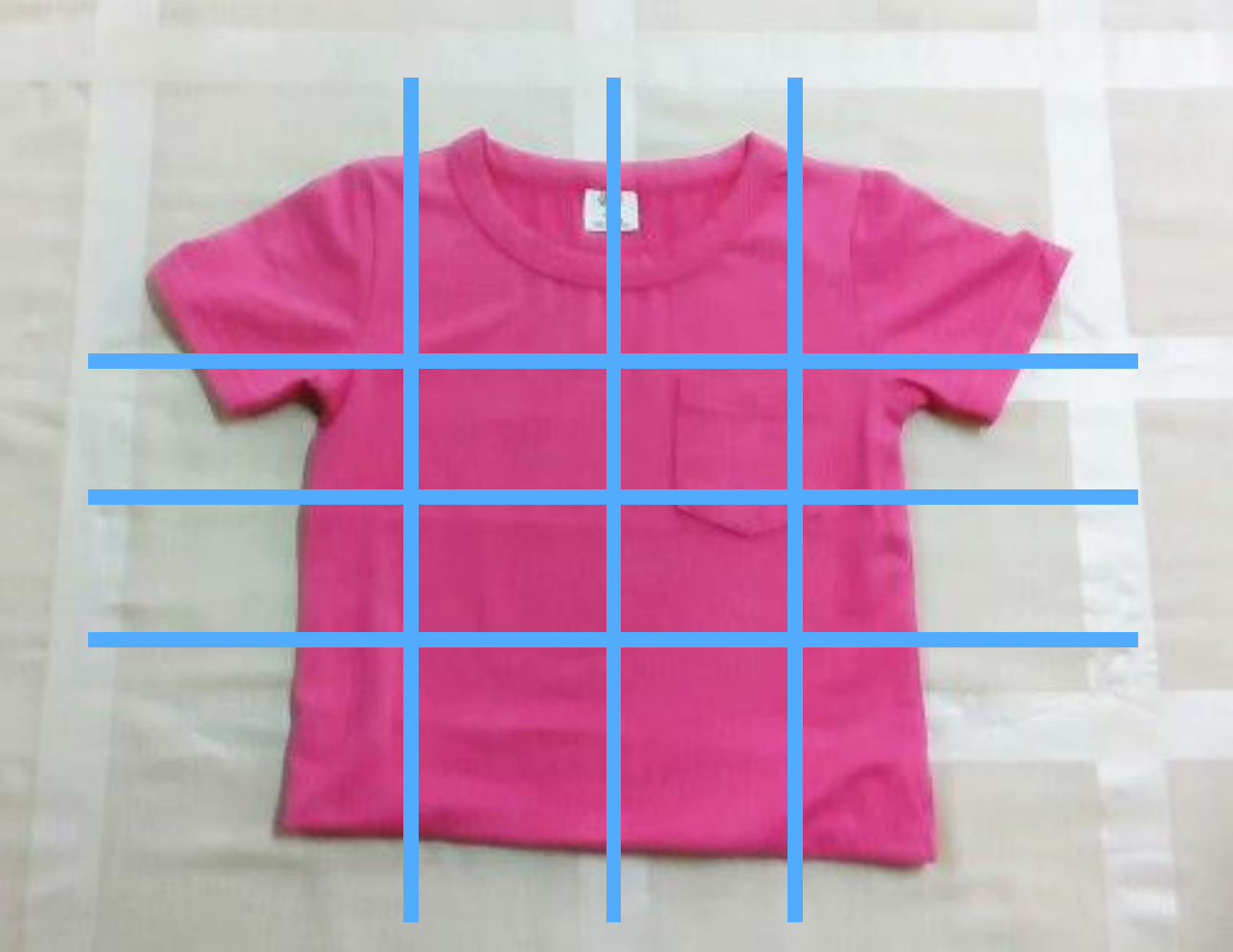}}
          \end{minipage}
          \vspace{0.25cm} \\
          Initial state & Shirts/Shorts are initialized to a state spread by a human. \vspace{0.25cm}\\ 
          Demonstrations & Sample is collected from humans for trajectory of $5$ episodes (shirt: $5$ episodes $\times 3$ steps $= 15$ samples, shorts: $5$ episodes $\times 2$ steps $= 10$ samples).\vspace{0.25cm}\\ 
          Goal state & Operator selected goal state from demonstration trajectories\\
          \bottomrule
          \end{tabular}
        }

        \subfloat[Parameter setting of GA-GAIL algorithm\label{table:fold_GAIL_setting}]{
          \begin{tabular}{@{}lp{8cm}llll@{}}
          \toprule
           \textbf{Parameter} & \textbf{Meaning} & \textbf{Value}  \\ \midrule
           $\eta$ & Parameters that control effect of smooth policy update & 0.7 \vspace{0.25cm} \\
           $\sigma$ & Parameters that control effect of causal entropy & 0.03 \vspace{0.25cm} \\
           $M$ & Number of episodes for one iteration & 10 \vspace{0.25cm} \\
           $T$ & Number of steps for one episode & \begin{tabular}{l}\hspace{-0.31cm} 3 (shirt) or \\\hspace{-0.31cm} 2 (shorts)\end{tabular} \vspace{0.25cm} \\
           $J$ & Iterations of discriminator updates & 10 \vspace{0.25cm} \\
           $K$ & Iterations of value network updates & 20 \vspace{0.25cm} \\ \bottomrule
          \end{tabular}
        }
    \end{center}
    \caption{Settings and learning parameters of folding clothes task\label{table:fold_learning_setting}}
\end{table}

\subsubsection{Results}
Since designing a reward function for this folding task is challenging, we evaluated the learned policies through learned trajectories, task success rates, and a Convolutional Neural Network (CNN) trained feature extraction of the policies. GA-GAIL learned folding policies from $390$ samples of shirts collected in about five hours and $220$ samples of shorts collected in about three hours. 

\begin{figure}
  \centering
  \subfloat[Shirt-folding trajectory\label{fig:shirt_trojectory}]{
      \includegraphics[width=0.6\columnwidth]{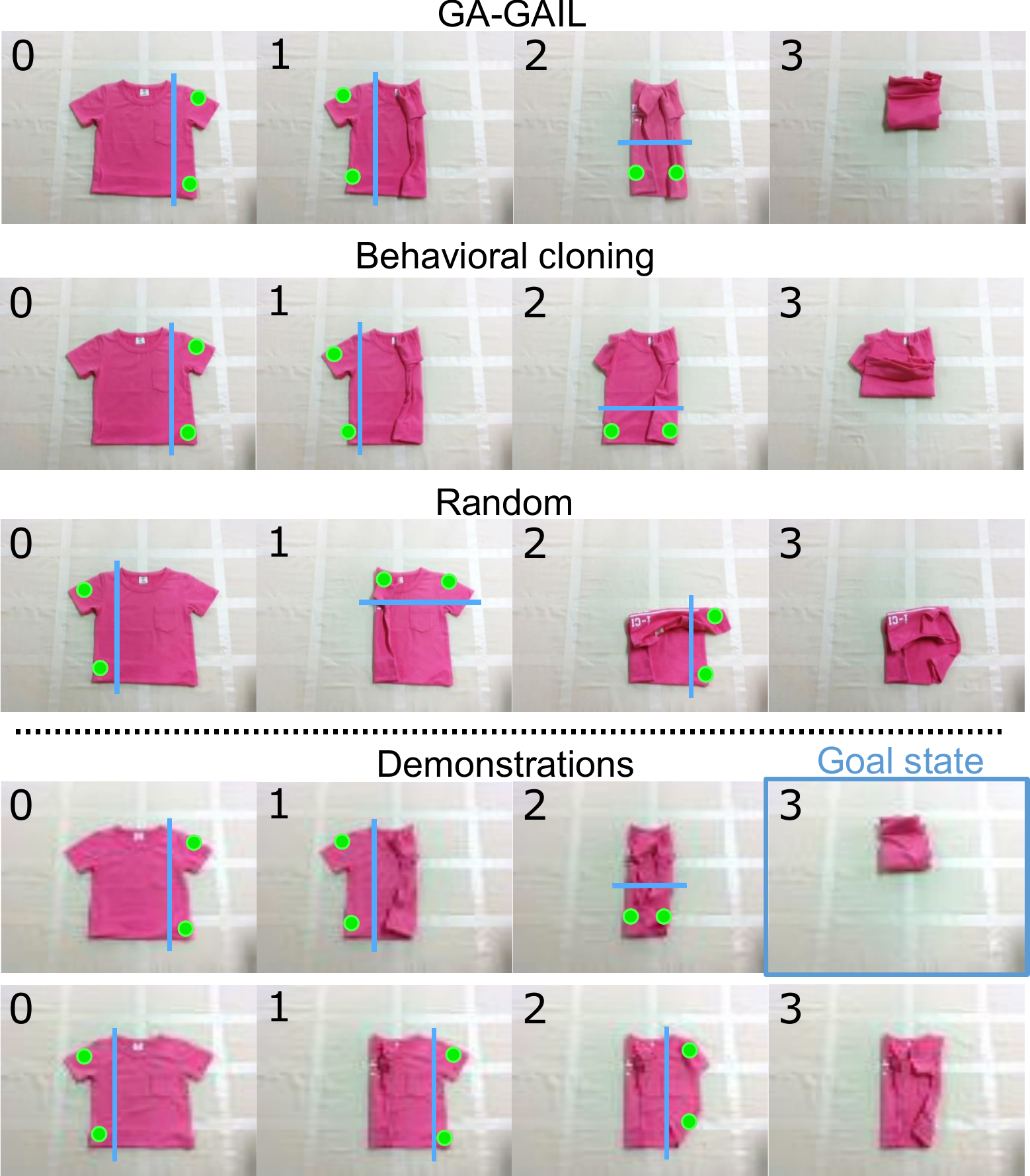}
  }
  \hspace{3mm}
  \subfloat[Shorts-folding trajectory\label{fig:trousers_trojectory}]{
      \includegraphics[width=0.5\columnwidth]{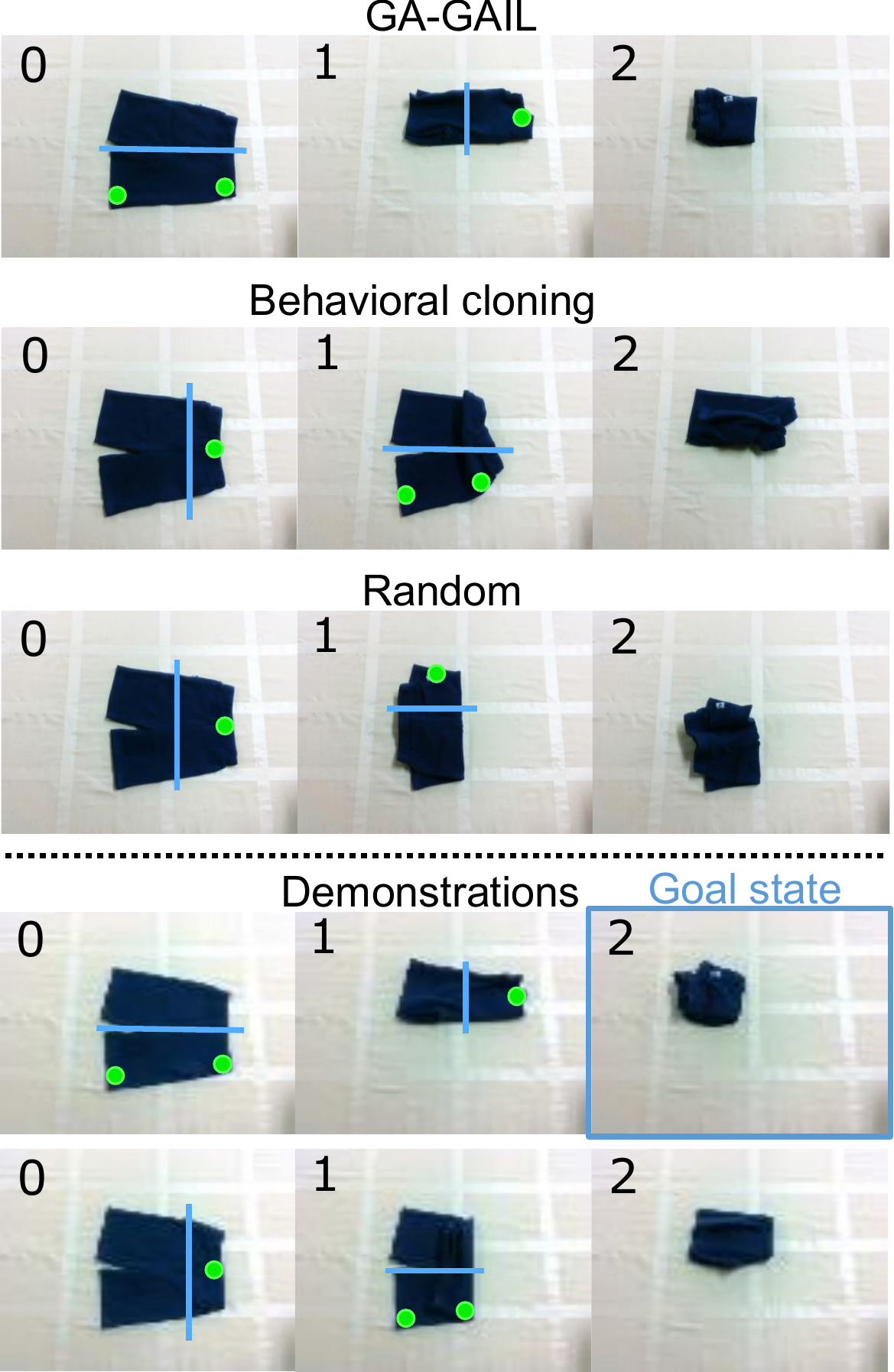}
  }
  \caption{Folding trajectory generated from learned policy: Blue frames in demonstrations show labeled goal state.}
  \label{fig:folding_trajectory}
\end{figure}

Figure \ref{fig:folding_trajectory} shows the trajectories generated from the policies learned for the two clothing-folding tasks. GA-GAIL's learning policies have a high success rate for both shirts and shorts, while the BC learning policies have a low success rate due to imperfect demonstrations. Table \ref{table:folding_init_test} shows the evaluation results of the success rate for each initial state. GA-GAIL policies have a higher success rate at different initial states, indicating that they are more robust than the BC policies.

\begin{table}
  \centering
  \begin{tabular}{cC{1.8cm}C{1.5cm}C{1.5cm}}
    \toprule
      \textbf{Initial state} & \textbf{GA-GAIL} & \textbf{BC} & \begin{tabular}{c}\textbf{\!\!\!\!\!Imperfect}\\\textbf{\!\!\!\!\!demo}\end{tabular} \\ \midrule
      \begin{minipage}{2cm}
      \centering
      \scalebox{0.8}{\includegraphics[width=1\columnwidth]{./fig/real_robo/fold/init_state/shirt_init0.png}}
      \end{minipage}
      & {\Large $\mathbf{4/5}$} & {\Large $1/5$} & {\Large $3/5$} \vspace{0.1cm} \\
      \begin{minipage}{2cm}
      \centering
      \scalebox{0.8}{\includegraphics[width=1\columnwidth]{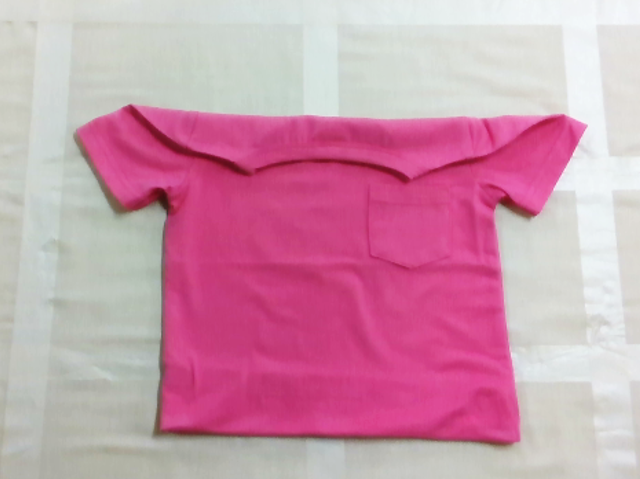}}
      \end{minipage}
      & {\Large $\mathbf{4/5}$} & {\Large $0/5$} & {\Large ---} \vspace{0.1cm} \\
      \begin{minipage}{2cm}
      \centering
      \scalebox{0.8}{\includegraphics[width=1\columnwidth]{./fig/real_robo/fold/init_state/trousers_init0.png}}
      \end{minipage}
      & {\Large $\mathbf{5/5}$} & {\Large $3/5$} & {\Large $3/5$} \vspace{0.1cm} \\
      \begin{minipage}{2cm}
      \centering
      \scalebox{0.8}{\includegraphics[width=1\columnwidth]{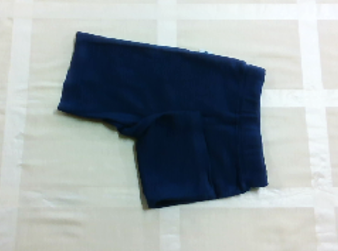}}
      \end{minipage}
      & {\Large $\mathbf{4/5}$} & {\Large $2/5$} & {\Large ---} \\ \bottomrule
  \end{tabular}
 \caption{Evaluation of success rate in clothing-folding task: Each method is evaluated with $5$ trials in each initial state. Success is defined as achieving goal state. The imperfect demonstration success rate was calculated using the demonstration trajectories used for learning.}\label{table:folding_init_test}
\end{table}

\begin{figure}
  \begin{center}
    \includegraphics[clip,scale=1.]{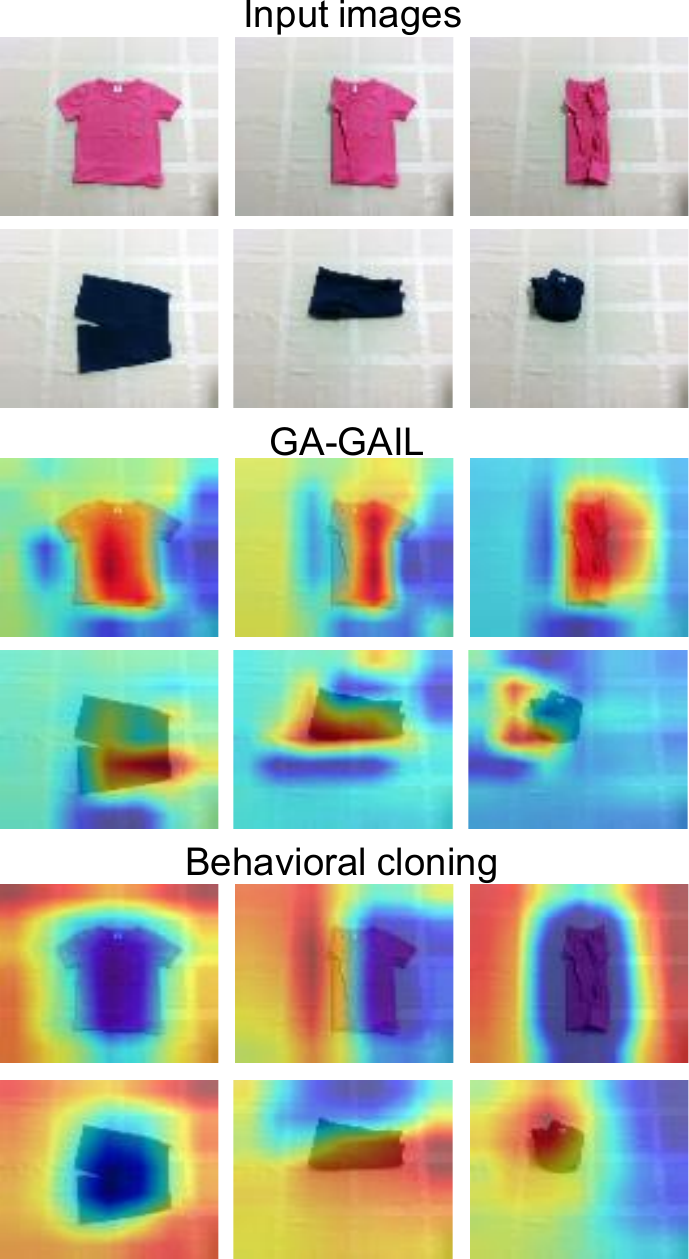}
    \caption{Visualization of extracted parts in images for action selection using Grad-CAM: Heat map shows parts extracted when actions are selected.}
    \label{fig:folding_grad_cam}
  \end{center}
\end{figure}

Grad-CAM \cite{grad_cam} visualized several examples of high-level features learned by GA-GAIL in Fig. \ref{fig:folding_grad_cam}. These heat maps, where red/blue colors indicate the high/low attention of the agent, indicate that our proposed method successfully learned useful and meaningful features. GA-GAIL policies extract such details as spreading areas and the edges of clothing, and the BC policies extract areas other than clothes. This result indicates that GA-GAIL policies are robust to position, wrinkles, and some creases.

\section{Discussion}\label{sec:dis}
In simulations and real-robot experiments, GA-GAIL learned policies with which it outperformed imperfect demonstrations. Our proposed method improved the task success rate of the learning policies by using goal labels collected at low cost from humans.

One main future challenge is automatically designing the action space. The proposed method learned policies without designing a task-specific state space and reward function, even though the learning environment requires the design of an action space for each task. We must determine what kind of action space is appropriate for learning environment settings. In real experiments of this research, cloth-manipulation tasks were learned using a high-level discrete action space, such as grasp-release points and folding lines. These action spaces contain hand-engineered, low-level policies that decide reaching and folding trajectories from observed images. Thus, the automatic design of action spaces requires learning low-level policies that are suitable for such tasks. When learning both high and low-level policies, the learning policy is structured in a hierarchy. In previous studies on learning hierarchical policy, hierarchical reinforcement learning obtains both policies to maximize the total reward \cite{NIPS2016_6233,DBLP:conf/icml/VezhnevetsOSHJS17,NIPS2018_7591}. Other hierarchical policies learn to imitate high-level policies from demonstration \cite{pmlr-v80-le18a}. 

Since the adversarial imitation learning framework imitates policy by trial and error, the learning cost of the real-robot environment is high. To lower this learning cost, the adversarial imitation learning framework for real-robot tasks requires efficient state initialization methods and sample efficient algorithms. Previous studies for state initialization methods have learned initialization policies at the same time as learning policies: a learning initialization policy seeks a state with low observation density \cite{Zhu2020The}; an initialization policy learns to achieve a specific initial state \cite{eysenbach2018leave}. Several previous works improved the sample efficiency of adversarial imitation learning: an agent of adversarial imitation learning estimated the state-action value without a learning discriminator \cite{sasaki2018sample}; a model-based adversarial imitation learning imitated learned state transition models \cite{schroecker2018generative}.

We believe that the theoretical analysis of GA-GAIL is essential to further clarify the differences from other GAIL frameworks. The GA-GAIL reward function consists of two parts: a goal reward, which can be formulated as a sparse reward function \cite{DBLP:conf/nips/AndrychowiczCRS17}, and a demonstration reward, which can be formulated as a comparatively dense reward function. By combining such reward function formulations with previous theoretical work \cite{DBLP:conf/nips/AndrychowiczCRS17} on GAIL, perhaps GA-GAIL can be analyzed theoretically by following previous studies \cite{pmlr-v119-zhang20d}. 

\section{Conclusion}\label{sec:con}
The contributions of this paper are twofold. The algorithmic aspect is GA-GAIL's proposal, a new adversarial imitation learning framework with two discriminators. 
GA-GAIL used a second discriminator, a goal label discriminator, to reduce the negative effects of imperfect demonstrations and learn policies that outperform imperfect demonstrations. GA-GAIL stabilized the policy updates with EDPN because the reward function represented by the two discriminators in the proposed method tends to change dynamically with two discriminator updates. EDPN contains a constraint that maximizes the policy's entropy with smooth policy updates, thus encouraging searching while reducing policy over-learning. 
This approach's application contribution is that it learned a cloth-manipulation policy without any specific reward function design. We applied our proposed method to a real-robot clothing manipulation task that consisted of a dual-armed robot called NEXTAGE and learned policies with high success rates from human demonstrations.

\section*{Acknowledgment}
This work was supported by JSPS KAKENHI Grant Number 21H03522 and JSPS Research Fellow Grant Number 20J11948.

\bibliography{main}

\end{document}